# Stochastic Optimization of Linear Dynamic Systems with Parametric Uncertainties

*Vadim Yatsenko*

This paper describes a new approach to solving some stochastic optimization problems for linear dynamic system with various parametric uncertainties. Proposed approach is based on application of tensor formalism for creation the mathematical model of parametric uncertainties. Within proposed approach following problems are considered: prediction, data processing and optimal control. Outcomes of carried out simulation are used as illustration of properties and effectiveness of proposed methods.

Various uncertainties are inherent to many control and/or sequential decision processes that come from technical, economical, biological applications etc. These uncertainties do not allow evaluate influences of all factors and, as a corollary, to apply a theory of deterministic system. Nature of uncertainties may be described as some stochastic processes, and theory of stochastic optimization must be used for analysis of considered system and synthesis of controls.

Main purpose of this paper is to provide a brief description of a new approach for solving some stochastic optimization problems for Linear Dynamic System (LDS) with Parametric Uncertainties (PUs) of various types. These solutions should be useful for various applications.

Overwhelming majority of existing stochastic optimization methods of LDS with PUs applies someway mapping of system states and inexact known parameters in somehow extended space. However in this case mathematical model of LDS will be transformed in non-linear system, that will makes difficulties for deriving the analytical solution and carrying out analyses of this ones.

For saving the method (structure) of interaction of LDS states and parameters it is proposed to use alternative approach. The basic idea of this approach is to apply tensor formalism for construction a mathematical description of LDS with PUs. As far as the random matrixes are applied for definition of LDS PUs, it is necessary to determine its statistical characteristics, namely expectations and covariances (second central moment). By analogy with well-known definitions of multidimensional statistical analysis - expectation of random vector is vector and covariance is matrix - expectation of random matrix will be certain matrix, and its covariance will be some mathematical object with four indexes, or tensor.

This idea will be discussed at inception of this paper. There will be introduced the definitions for tensors which specifies expectation, second central moment of random square matrixes and covariance of random matrixes and vectors. Inasmuch as tensor analyses do not operate with mathematical objects similar to rectangular matrixes, there will be introduced conception of dual tensors, i.e. tensors that are defined in two different metric domains (spaces). Further there will be introduced definitions for expectation and second central moments of random dual tensors, covariance of random dual tensors and vectors, and covariance of random dual tensors and ordinary tensors (square matrixes) as well. At the end of this section there will be introduced definitions for reducible and irreducible PUs. Second part deals with the problems of data processing in LDS with PUs. The involved problems are: prediction, filtration and identification for LDS with PUs of various types. Futher new problem will be considered - informational prediction of solving the identification problem. In final part will be treated the optimal control problems for LDS with PUs. There will be derived the formulas for optimal controls for LDS with irreducible and reducible PUs. At that there will be demonstrated that proposed optimal controls for LDS with reducible PUs have all properties of dual controls. In what follows there will be outlined an approach to solution of a new problem – optimal control for experimental development of LDS with reducible PU. Consideration will be finished by presentation of simulation outcomes that were carried out for overall proposed data processing methods and optimal controls.

## Definitions

### *Linear Dynamic Systems*

In the context of this consideration LDS term implies mathematical objects that are described by discrete-time equations

$$x(k+1) = F(k)x(k) + G(k)u(k) + w(k) \qquad (1)$$
$$z(k) = H(k)x(k) + v(k) \qquad (2)$$

where: $x(.) \in R_X^N$ vector of system states; $u(.) \in R_U^M$ - vector of controls; $F(.)$ and $G(.)$ – matrixes of inexact known parameters of corresponding dimensions; $w(.) \in R_X^N$ - vector of input noises; $z(.) \in R_Z^L$ - vector of measurements (observations); $H(.)$ – matrix of parameters of applied measuring system; $v(.) \in R_Z^L$ - vector of measurement interferences. Noises $w(k)$ and $v(k)$ are considered as independent Gaussian random processes with zero mean and $var(w(k))=Q(k)$, $var(v(k))=R(k)$ where $Q(.)$ - positive semidefinite, and $R(.)$ - positive definite symmetrical matrixes.

### *Random Matrixes, its Moments, Covariance of Random Vectors and Matrixes*

In what follows everywhere will be assumed that $\{x(.),F(.),G(.)\}$ set of system states and inexact known parameters of system are distributed in accordance with Gaussian law. It is known that Gaussian distribution of some random set is uniquely defined by its expectation and second central moment (dispersion matrix). Therefore, if there will be stated mathematical objects that unambiguously define expectation and second central moment of $\{x(.),F(.),G(.)\}$ random set, then will be defined the Gaussian distribution of this one.

As mentioned above, for solving considered problems will be applied a new approach which is based on usage of tensor calculus [4,5], and this fact requires providing of some additional treatments. Below will be introduced formal definitions for set of mathematical objects, which will define expectation and second central moment of $\{x(.),F(.),G(.)\}$ set.

Let there is defined Riemannian space $R_X^N$, $x(.) \in R_X^N$ with metric tensor $g_{ij}$ (associated tensor $g^{ij}$)[1]. Then it is possible to introduce following definitions[2].

*Definition 1.* $x(.)$ vector (covector) is referred to as random one, if its components are random values.

*Definition 2.* $\bar{x}(.)$ vector (covector) is defined as expectation of $x(.)$ random vector (covector), if its components are related by

$$\bar{x}^i(.) = M\{x^i(.)\}, \; (\bar{x}_i(.) = M\{x_i(.)\})$$

---
[1] Further will be considered spaces with Euclidean metrics, i.e. $g(x)_{ij}=\delta_{ij}$ ($g^{ij}(x)=\delta^{ij}$), where $\delta_{ij}$ – Kronecker delta. However derived relations may be theorized for case of homogeneous spaces, i.e. for case of $g_{ij}(x)=g_{ij}$ ($g^{ij}(x)=g^{ij}$).
[2] Following five definitions are formal ones, and properly are well-known definitions of classic statistical analyses, which are redefined in terms of tensor analyses.

*Definition 3.* $\alpha(.)$ matrix (tensor of rang 2, type (1,1)) is referred to as second central moment random vector $x(.)$, if its components are related by

$$\alpha_j^i(.) = \mathrm{var}\{x^i(.)\} = M\{(x^i(.) - \bar{x}^i(.))g_{jk}(x^k(.) - \bar{x}^k(.))\}$$

In case of $x_j(.)$ covector its covariance matrix may be defined as

$$\tilde{\alpha}_j^i(.) = \mathrm{var}\{x_j(.)\} = M\{(x_j(.) - \bar{x}_j(.))g^{ik}(x_k(.) - \bar{x}_k(.))\}$$

It is obviously that $\alpha(.) = \tilde{\alpha}(.)$.

*Definition 4.* $F(.)$ square matrix (tensor of rang 2, type (1,1)) is referred to as random one, if its elements are random values[3].

*Definition 5.* $\bar{F}(.)$ matrix (tensor of rang 2, type (1,1)) is defined as expectation of random matrix $F(.)$, if its components are related by

$$\bar{F}_j^i(.) = M\{F_j^i(.)\}.$$

Let there in $R_X^N$ space is defined $F(.)$ random matrix with $\bar{F}(.)$ expectation and $x(.)$ random vector with $\bar{x}(.)$ expectation.

*Definition 6.* Second central moment of $F(.)$ random matrix will be $\beta(.)$ tensor of rang 4, type (2,2), which components will be defined as

$$\beta_{kl}^{ij}(.) = \{\mathrm{var}(F(.))\}_{kl}^{ij} = M\{(F_k^i(.) - \bar{F}_k^i(.))g_{ml}g^{nj}(F_n^m(.) - \bar{F}_n^m(.))\}$$

*Definition 7.* First cross moment of $F(.)$ random matrix and $x(.)$ random vector will be $\varphi(.)$ tensor of rang 3, type (1,2), which components will be defined as

$$\varphi_{jk}^i(.) = \{\mathrm{cov}(F(.), x(.))\}_{jk}^i = M\{(F_j^i(.) - \bar{F}_j^i(.))g_{mk}(x^m(.) - \bar{x}^m(.))\}$$

*Definition 8.* First cross moment of $x(.)$ random vector and $F(.)$ random matrix will be $\bar{\varphi}(.)$ tensor of rang 3, type (2,1), which components will be defined as

$$\bar{\varphi}_k^{ij}(.) = \{\mathrm{cov}(x(.), F(.))\}_k^{ij} = M\{(x^i(.) - \bar{x}^i(.))g_{mk}g^{nj}(F_n^m(.) - \bar{F}_n^m(.))\}$$

It is easy to see that $\varphi(.)$ and $\bar{\varphi}(.)$ tensors will be related by $\varphi_{jk}^i(.) = g^{il}g_{km}g_{jn}\bar{\varphi}_l^{mn}(.)$

Within tensor analyses are not defined the mathematical objects similar to rectangular matrixes. This fact causes a necessity to introduce following concept. Let there are defined two Riemannian spaces $R_X^N$, $x(.) \in R_X^N$ and $R_U^M$, $u(.) \in R_U^M$ with metric tensors $g_{ij}$ и $h_{ij}$ respectively (associated tensors $g^{ij}$ и $h^{ij}$). $x(.)$ mathematical object, which is defined as vector with respect to $R_X^N$, may be considered as some scalar with respect to $R_U^M$ since $x(.)$ is not changed when $R_U^M$ space will be transformed. And vice versa, $u(.)$ mathematical object, which is defined as vector in $R_U^M$, with respect to $R_X^N$ may be considered as some scalar. Let $x(.)$ and $u(.)$ are related by $x(.)=G(.)u(.)$ where $G(.)$- some rectangular matrix. Then for all $T_x$ and $T_u$ transformations of $R_X^N$ и $R_U^M$ spaces the following equation will be true

$$T_x x(.) = T_x G(.) T_u^{-1} T_u u(.)$$

---
[3] It is not necessary that all elements of matrix will be random values; proposed approach includes the case when some parameters of model are exact known, i.e. when parameters set obey the degenerate Gaussian law.

It is therefore concluded that $G(.)$ matrix may be considered as some dual mathematical object, which is vector with respect to $R_X^N$ space (tensor of rang 1, type (1,0)), and covector with respect to $R_U^M$ space (tensor of rang 1, type (0,1)).

Further consideration will lead to necessity of introduction the more complex dual tensors, for definition of which will be applied following notation: tensor of rang $(n_x, n_u)$, type $((k_x, l_x), (k_u, l_u))$, where $n_x$ and $(k_x, l_x)$ are rang and type of tensor with respect to $R_X^N$ space; $n_u$ and $(k_u, l_u)$ are rang and type of tensor with respect to $R_U^M$ space. Then it is possible to introduce following definitions – analogues of (4)-(8) – for random dual tensors.

*Definition 9.* $G(.)$ rectangular matrix (dual tensor of $R_X^N$ и $R_U^M$ of rang (1,1), type ((1,0),(0,1))) will be referred to as random one, if its elements are random values.

*Definition 10.* $\overline{G}(.)$ rectangular matrix (dual tensor of $R_X^N$ и $R_U^M$ of rang (1,1), type ((1,0),(0,1))) will be referred to as expectation of $G(.)$ random matrix, if its components are related by

$$\overline{G}_j^i(.) = M\{G_j^i(.)\}$$

*Definition 11.* Second central moment of $G(.)$ random dual tensor of rang (1,1), type ((1,0),(0,1)) will be referred to as $\gamma(.)$ dual tensor of rang (2,2), type ((1,1),(1,1)), which components will be defined by

$$\gamma_{kl}^{ij}(.) = \{\mathrm{var}(G(.))\}_{kl}^{ij} = M\{(G_k^i(.) - \overline{G}_k^i(.))g_{lm}h^{jn}(G_n^m(.) - \overline{G}_n^m(.))\}$$

*Definition 12.* First cross moment of $G(.)$ dual tensor and $x(.)$ random vector will be defined to as $\psi(.)$ dual tensor of rang (2,1), type ((1,1),(0,1)), which components will be defined by

$$\psi_{jk}^i(.) = \{\mathrm{cov}(G(.),x(.))\}_{jk}^i = M\{(G_j^i(.) - \overline{G}_j^i(.))g_{mk}(x^m(.) - \overline{x}^m(.))\}$$

*Definition 13.* First cross moment of $x(.)$ random vector and $G(.)$ random dual tensor will be referred $\overline{\psi}(.)$ dual tensor of rang (2,1), type ((1,1),(1,0)), which components will be defined by

$$\overline{\psi}_k^{ij}(.) = \{\mathrm{cov}(x(.),G(.))\}_k^{ij} = M\{(x^i(.) - \overline{x}^i(.))g_{mk}h^{nj}(G_n^m(.) - \overline{G}_n^m(.))\}$$

It may be shown, that $\psi(.)$ and $\overline{\psi}(.)$ dual tensors are related by

$$\psi_{jk}^i(.) = g^{il}h_{jn}g_{km}\overline{\psi}_l^{mn}(.)$$

*Definition 14.* First cross moment of $F(.)$ random square and $G(.)$ rectangular matrixes and will be defined as

$$v_{kl}^{ij}(.) = \{\operatorname{cov}(F(.), G(.))\}_{kl}^{ij} = M\{(F_k^i(.) - \overline{F}_k^i(.))g_{ml}h^{nj}(G_n^m(.) - \overline{G}_n^m(.))\}$$

From this follows that $v(.)$ will be dual tensor of rang (3,1), type ((1,2),(1,0)).

Analogously may be defined $\overline{v}(.)$ first cross moment of rectangular and square matrixes that must be a dual tensor of rang (3,1), type ((2,1),(0,1)).

Thus above are introduced all definitions for expectations, first cross moments and second central moments for the $\{x(.), F(.), G(.)\}$ random set. It should be pointed out once more that $P(.) = \{\overline{x}(.), \overline{F}(.), \overline{G}(.), \alpha(.), \beta(.), \gamma(.), v(.), \varphi(.), \psi(.)\}$ set unambiguously define Gaussian distribution of $\{x(.), F(.), G(.)\}$ random set. Hereafter the following notations will be used

$$\overline{x}(.) = M\{x(.)\};\ \overline{F}(.) = M\{F(.)\};\ \overline{G}(.) = M\{G(.)\}$$
$$\alpha(.) = \operatorname{var}\{x(.))\};\beta(.)\} = \operatorname{var}\{F(.))\};\gamma(.)\} = \operatorname{var}(G(.));v(.) = \operatorname{cov}\{F(.), G(.)\};$$
$$\varphi(.) = \operatorname{cov}\{F(.), x(.)\}; \psi(.) = \operatorname{cov}\{G(.), x(.)\}$$

In addition to this below will be applied following designations for conjugated first cross moments

$$\overline{v}(.) = \operatorname{cov}\{G(.), F(.)\};\ \overline{\varphi}(.) = \operatorname{cov}\{x(.), F(.)\};\ \overline{\psi}(.) = \operatorname{cov}\{x(.), G(.)\}$$

Notice that introduced mathematical objects (tensors, dual tensors) possess an invariant property with respect to all permissible transformation of $R_X^N$ and $R_U^M$ spaces. From this results that correct equations, which will be constructed with application of these objects, will have the invariant property with respect to above transformations.

### Models of PUs

In the context of this consideration PU term implies that $\{F(.), G(.)\}$ matrixes of system parameters are known inexact (all elements or some subset). At the same time $H(.)$ matrix is assumed as exact known a priory – majority of real systems obeys this assumption. Provided below treatment may be theorized for case of uncertainty in elements of $H(.)$ matrix, if need arise [4].

In what follows two types of PUs to be considered – irreducible and reducible. Irreducible PU is characterized by influence on system parameters unpredictable and/or random factors, which may be described by mathematical model of fluctuation, or stochastic walks. In this case unknown parameters will be considered as samples from some stochastic process such as white noise with a priory determined statistical characteristics, i.e. by $\{\overline{F}(.), \overline{G}(.), \beta(.), \gamma(.), v(.)\}$ set. As this takes place, this PU model assumes that estimates of system states and parameters are uncorrelated. Information state for LDS with irreducible PU will be $P_I(.) = \{\overline{x}(.), \alpha(.)\}$ set

---

[4] In this case it is necessary to introduce "triple" tensors, or tensor that are defined in $R_x, R_u$ и $R_z$ spaces. In this case there is need to define corresponding random tensors, its expectations, second central moments, etc, i.e. full statistical description for $\{x(.), F(.), G(.), H(.)\}$ random set. Then it is possible (by analogy with provided below consideration) to derive the relations for systems with PUs of such type.

As reducible PU will be considered the uncertainty, which may be reduced by experimental development of system, i.e. by testing and follow-up processing of measured data. This type of PU is characterized by a priory lack of knowledge about true values of parameters. As this takes place, it is important to take into account the fact that parameters take some predetermined values. This fact may be expressed by following equations

$$F(k+1) = F(k) \tag{3}$$
$$G(k+1) = G(k) \tag{4}$$

When will be considered LDS with reducible PU, there will be applied information state of type $P_R(.) = \{\bar{x}(.), \bar{F}(.), \bar{G}(.), \alpha(.), \beta(.), \gamma(.), \nu(.), \varphi(.), \psi(.)\}$. In addition both PU models suppose that $\{F(k), G(k)\}$ set and $w(k)$, $v(k)$ noises are uncorrelated.

## Data Processing in LDS with PUs

### *Prediction for Autonomous LDS with PUs*

Autonomous LDS is described by following equation

$$x(k+1) = F(k)x(k) + w(k) \tag{5}$$

Initial information state for prediction of LDS with irreducible PU is $P_I(k) = \{\bar{x}(k), \alpha(k)\}$. (5) in effect defines the mapping of $\{x(k), F(k), w(k)\}$ set in $x(k+1)$ random vector. It is easy to derive following relations using above assumptions and definitions

$$\bar{x}(k+1|k) = \bar{x}(k+1|P_I(k)) = M\{x(k+1)|P_I(k)\} = M\{F(k)x(k) + w(k)|P_I(k)\} =$$
$$= M\{(\bar{F}(k) + \Delta F(k))(\bar{x}(k) + \Delta x(k)) + w(k)|P_I(k)\} =$$
$$= M\{\bar{F}(k)\bar{x}(k) + \Delta F(k)\bar{x}(k) + \bar{F}(k)\Delta x(k) + \Delta F(k)\Delta x(k) + w(k)|P_I(k)\} = \bar{F}(k)\bar{x}(k)$$

$$\alpha(k+1|k) = \mathrm{var}(x(k+1)|P_I(k)) = M\{(x^i(k+1) - \bar{x}^i(k+1))g_{jk}(x^k(k+1) - \bar{x}^k(k+1))|P_I(k)\} =$$
$$M\{((\bar{F}(k) + \Delta F(k))^i_m(\bar{x}(k) + \Delta x(k))^m + w^i(k) - \bar{F}^i_m(k)\bar{x}^m(k))g_{jk}((\bar{F}(k) + \Delta F(k))^k_n(\bar{x}(k) + \Delta x(k))^n +$$
$$+ w^k(k) - \bar{F}^k_n(k)\bar{x}^n(k))|P_I(k)\} = M\{(\bar{F}^i_m(k)\Delta x^m + \Delta F^i_m(k)\bar{x}^m(k) + \Delta F^i_m(k)\Delta x^m(k) + w^i(k))g_{jk}$$
$$(\bar{F}^k_n(k)\Delta x^n(k) + \Delta F^k_n(k)x^n(k) + \Delta F^k_n(k)\Delta x^n(k) + w^k(k))|P_I(k)\}$$

Taking into account that all third moments of Gaussian distributed set are equal to zero, as well as assumption about non-correlatedness of $w(k)$ and members of $\{x(k), F(k)\}$ set, it may be obtained the following[5]

$$\alpha(k+1|k) = M\{\bar{F}^i_m(k)\Delta x^m(k)g_{jk}\bar{F}^k_n(k)\Delta x^n(k) + \Delta F^i_m(k)\bar{x}^m(k)g_{jk}\Delta F^k_n(k)\bar{x}^n(k) +$$
$$+ \Delta F^i_m(k)\Delta x^m(k)g_{jk}\Delta F^k_n(k)\Delta x^n(k) + w^i(k)g_{jk}w^k(k)|P_I(k)\} =$$

---

[5] Hereafter for decreasing the volume of formulas recording, along with tensor notation will be used matrix one. At that will be applied embedded form of notation – if the main part of formula applies tensor form, then braces point out it parts, which is recorded with application of matrix notation. And vice versa: if formula record uses matrix notation, then braces mark out the part that is recorded with usage of tensor notation. Sometimes will be used nested forms of this rule.

$$M\{\bar{F}_m^i(k)\Delta x^m(k)g_{jk}\bar{F}_n^k g^{np}g_{pq}(k)\Delta x^q(k)\,|\,P_I(k)\}+$$

$$+M\{\Delta F_m^i(k)\bar{x}^m(k)g_{jk}\Delta F_n^k g^{np}g_{pq}(k)\bar{x}^q(k)\,|\,P_I(k)\}+$$

$$+M\{\Delta F_m^i(k)\Delta x^m(k)g_{jk}\Delta F_n^k(k)g^{np}g_{pq}(k)\Delta x^q(k)\,|\,P_I(k)\}+M\{w^i(k)g_{jk}w^k(k)\,|\,P_I(k)\}=$$

$$=\bar{F}_m^i(k)M\{\Delta x^m(k)g_{pq}\Delta x^q(k)\,|\,P_I(k)\}g_{jk}g^{np}\bar{F}_n^k(k)+M\{\Delta F_m^i(k)g_{jk}g^{np}\Delta F_n^k\,|\,P_I(k)\}\bar{x}^m(k)\bar{x}_p(k)+$$

$$+M\{\Delta F_m^i(k)\Delta x^m(k)g_{jk}\Delta F_n^k(k)g^{np}g_{pq}\Delta x^q(k)\,|\,P_I(k)\}+Q(k)=$$

$$=\bar{F}(k)\alpha(k)\bar{F}^T(k)+Q(k)+\{\beta_{mj}^{ip}(k)\bar{x}^m(k)\bar{x}_p(k)+\beta_{mj}^{ip}(k)\alpha_p^m(k)\}$$

Last formula uses following designations $\bar{x}_p(k)=g_{pq}\bar{x}^q(k)$, $\{\bar{F}^T(k)\}_j^i=g_{jp}g^{iq}\bar{F}_q^p(k)$ - these are well-known transposition operation of matrix analyses, and tensor analog of relationship for fourth central moment of $X$ Gaussian random vector with $\sigma_{ij}$ covariance matrix [1][6] $M(\Delta X_i \Delta X_j \Delta X_k \Delta X_l)=\sigma_{ij}\sigma_{kl}+\sigma_{ik}\sigma_{jl}+\sigma_{il}\sigma_{jk}$ which for considered case is given by

$$M\{\Delta F_m^i(k)\Delta x^m(k)g_{jk}\Delta F_n^k(k)g^{np}g_{pq}\Delta x^q(k)\,|\,P_I(k)\}=\{\beta_{mj}^{ip}(k)\alpha_p^m(k)\}$$

Components of $P_I(k+1|k)=\{\bar{x}(k+1|k),\alpha(k+1|k)\}$ prediction are given by

$$\bar{x}(k+1|k)=M\{x(k+1)\,|\,P_I(k)\}=\bar{F}(k)\bar{x}(k)$$

$$\alpha(k+1|k)=\mathrm{var}(x(k+1)\,|\,P_I(k))=\bar{F}(k)\alpha(k)\bar{F}^T(k)+Q(k)+\{\beta_{lj}^{ik}(k)(\alpha_k^l(k)+\bar{x}^l(k)\bar{x}_k(k))\}$$

$P_R(k)=\{\bar{x}(k),\bar{F}(k),\alpha(k),\beta(k),\varphi(k)\}$ set determines information state of LDS with reducible PU. Applying derived relations and (3) it is simply to obtain similar formulas for components of $P_R(k+1|k)=\{\bar{x}(k+1|k),\bar{F}(k+1|k),\alpha(k+1|k),\beta(k+1|k),\varphi(k+1|k)\}$ set.

$$\bar{x}(k+1|k)=M\{x(k+1)\,|\,P_R(k)\}=\bar{F}(k)\bar{x}(k)+\{\varphi_{jk}^i(k)g^{jk}\}$$

$$\bar{F}(k+1|k)=M\{F(k+1)\,|\,P_R(k)\}=\bar{F}(k)$$

$$\alpha(k+1|k)=\mathrm{var}(x(k+1)\,|\,P_R(k))=\bar{F}(k)\alpha(k)\bar{F}^T(k)+Q(k)+A+A^T+$$

$$+\{\beta_{lj}^{ik}(k)[\alpha_k^l(k)+\bar{x}^l(k)\bar{x}_k(k)]+\varphi_{kl}^i(k)\bar{\varphi}_j^{kl}(k)\}$$

$$\beta_{jk}^{il}(k+1|k)=\{\mathrm{var}(F(k+1)\,|\,P_R(k)\}_{jk}^{il}=\beta_{jk}^{il}(k)$$

$$\varphi_{jk}^i(k+1|k)=\{\mathrm{cov}(F(k+1),x(k+1))\,|\,P_R(k)\}_{jk}^i=\bar{F}^T(k)_k^l \varphi_{jl}^i(k)+\beta_{jk}^{il}(k)\bar{x}_l(k)$$

where $A=\{\bar{F}_k^i(k)\bar{\varphi}_j^{kl}(k)\bar{x}_l(k)\}$

It is easy to see that formula for $\alpha(k+1|k)$ is similar to one of Kalman filter - difference consist in appearance of additional terms, which are proportional to $\beta(k)$. This fact is of essential significance – as far as $\beta(k)$ tensor may be considered as positively semidefinite (this statement uses matrix analogies once more), then, even in case of $\bar{x}(k)=0$ and stability of $\bar{F}(k)$ matrix, solutions for $\alpha(k+1|k)$ may be unstable, and neglecting of PU factor may leads to sizeable drifts of predicted estimates. This one may be some additional explanation for divergence of Kalman filter due to PU factor.

---

[6] This is first but not last application of heuristic analogies of well-known formulas of matrix calculus and statistical analysis for $\{x(.),F(.),G(.)\}$ random set.

## General Case of Prediction for LDS

Applying above results, after bulky but light calculation may be derived the following relations for components of $P_R(k+1|k)$ set

$$\bar{x}(k+1|k) = M\{x(k+1)|P_R(k)\} = \overline{F}(k)\bar{x}(k) + \overline{G}(k)u(k) + \{\varphi^i_{jk}(k)g^{jk}\}$$

$$\overline{F}(k+1|k) = M\{F(k+1)|P_R(k)\} = \overline{F}(k)$$

$$\overline{G}(k+1|k) = M\{G(k+1)|P_R(k)\} = \overline{G}(k)$$

$$\alpha(k+1|k) = \text{var}(x(k+1)|P_R(k)) = \overline{F}(k)\alpha(k)\overline{F}^T(k) + Q(k) + A + A^T +$$
$$+ \{\beta^{ik}_{lj}(k)(\alpha^l_k(k) + \bar{x}^l(k)\bar{x}_k(k)) + \gamma^{ik}_{lj}(k)u^l(k)u_k(k) + \varphi^i_{kl}(k)\overline{\varphi}^{kl}_j(k)\}$$

$$\beta^{il}_{jk}(k+1|k) = \{\text{var}(F(k+1)|P_R(k)\}^{il}_{jk} = \beta^{il}_{jk}(k)$$

$$\gamma^{ik}_{jl}(k+1|k) = \{\text{var}(G(k+1)|P_R(k)\}^{ik}_{jl} = \gamma^{ik}_{jl}(k)$$

$$v^{ik}_{jl}(k+1|k) = \{\text{cov}(F(k+1),G(k+1)|P_R(k)\}^{ik}_{jl} = v^{ik}_{jl}(k)$$

$$\varphi^i_{jk}(k+1|k) = \{\text{cov}(F(k+1),x(k+1)|P_R(k)\}^i_{jk} = \overline{F}^T(k)|^l_k \varphi^i_{jl}(k) + \beta^{il}_{jk}(k)\bar{x}_l(k) + v^{il}_{jk}(k)u_l(k)$$

$$\psi^i_{jk}(k+1|k) = \{\text{cov}(G(k+1),x(k+1)|P_R(k)\}^i_{jk} = \overline{F}^T(k)|^l_k \psi^i_{jl}(k) + v^{il}_{jk}(k)\bar{x}_l(k) + \gamma^{il}_{jk}(k)u_l(k)$$

Here are used the notations: $\bar{x}_i(k) = g_{ij}\bar{x}^j(k)$, $u_i(k) = h_{ij}u^j(k)$ (these ones will be applied below) and

$$A = \{v^{ik}_{lj}(k)\bar{x}^l(k)u_l(k) + \overline{F}^i_k(k)\overline{\varphi}^{kl}_j(k)\bar{x}_l(k) + \overline{F}^i_k(k)\overline{\psi}^{kl}_j(k)u_l(k)\}.$$

Components of $P_I(k+1|k)$ are given by

$$\bar{x}(k+1|k) = M\{x(k+1)|P_I(k)\} = \overline{F}(k)\bar{x}(k) + \overline{G}(k)u(k)$$

$$\alpha(k+1|k) = \text{var}(x(k)|P_I(k)) = \overline{F}(k)\alpha(k)\overline{F}^T(k) + Q(k) + A + A^T +$$
$$+ \{\beta^{ik}_{lj}(k)(\alpha^l_k(k) + \bar{x}^k(k)\bar{x}_l(k)) + \gamma^{ik}_{lj}(k)u^l(k)u_k(k)\}$$

here $A = \{v^{ik}_{lj}(k)\bar{x}^l(k)u_k(k)\}$

## Data Processing in LDS with PUs

Data processing algorithms for LDS with PUs will be derived without consideration of corresponding stochastic optimization problem as before. Equations for components of information state will be obtained with usage of well-known formulas for conditional normal distribution subject to applied measurements model (2) and derived relations for prediction of information states.

Let consider a join distribution of $x(k+1) \in R^N_X$ random vector that is distributed in accordance with Gaussian law with $\{\bar{x}(k+1), \alpha(k+1)\}$, and $z(k+1)$ random vector, which is determined by (2). Then $X(k+1) = \begin{pmatrix} x(k+1) \\ z(k+1) \end{pmatrix}$ random vector is normally distributed as well, and parameters of distribution will be $\overline{X}(k+1) = \begin{pmatrix} \bar{x}(k+1) \\ \bar{z}(k+1) \end{pmatrix}$ and $\text{var}\{X(k+1)\} = \begin{bmatrix} \alpha(k+1) & \text{cov}\{z(k+1),x(k+1)\} \\ \text{cov}\{x(k+1),z(k+1)\} & \text{var}\{z(k+1)\} \end{bmatrix}$ where

$$\bar{z}(k+1) = M\{H(k+1)x(k+1) + v(k+1)\} = H(k+1)\bar{x}(k+1) \tag{6}$$
$$\text{var}\{z(k+1)\} = \text{var}\{H(k+1)x(k+1) + v(k+1)\} = H(k+1)\alpha(k+1)H^T(k+1) + R(k+1) \tag{7}$$
$$\text{cov}\{z(k+1), x(k+1)\} = H(k+1)\alpha(k+1) \tag{8}$$
$$\text{cov}\{x(k+1), z(k+1)\} = \alpha(k+1)H^T(k+1) \tag{9}$$

Then, in accordance with [1], following relations must be true

$$\bar{x}(k+1 \mid z(k+1)) = \bar{x}(k+1) + \text{cov}\{x(k+1), z(k+1)\} \text{var}\{z(k+1)\}^{-1} (z(k+1) - \bar{z}(k+1)) \tag{10}$$
$$\alpha(k+1 \mid z(k+1)) = \alpha(k+1) - \text{cov}\{x(k+1), z(k+1)\} \text{var}\{z(k+1)\}^{-1} \text{cov}\{z(k+1), x(k+1)\} \tag{11}$$

Immediate substitution of (6)-(9) in (10) и (11) demonstrates, that derived relations are similar to Kalman filter

$$\bar{x}(k+1 \mid k+1) = \bar{x}(k+1 \mid k) + \alpha(k+1 \mid k)H^T(k+1)[H(k+1)\alpha(k+1 \mid k)H^T(k+1) +$$
$$+ R(k+1)]^{-1}(z(k+1) - H(k+1)\bar{x}(k+1 \mid k))$$
$$\alpha(k+1 \mid k+1) = \alpha(k+1 \mid k) - \alpha(k+1 \mid k)H^T(k+1)[H(k+1)\alpha(k+1 \mid k)H^T(k+1) +$$
$$+ R(k+1)]^{-1}H(k+1)\alpha(k+1 \mid k)$$

### *Filtration for LDS with irreducible PU*

For elements of $P_I(k+1 \mid k+1) = P_I(k+1 \mid k, z(k+1))$ information state hold true following relations

$$\bar{x}(k+1 \mid k+1) = \bar{x}(k+1 \mid k) +$$
$$+ \alpha(k+1 \mid k)H^T(k+1)(H(k+1)\alpha(k+1 \mid k)H^T(k+1) + R(k+1))^{-1}(z(k+1) - H(k+1)\bar{x}(k+1 \mid k))$$
$$\alpha(k+1 \mid k+1) = \alpha(k+1 \mid k) -$$
$$- \alpha(k+1 \mid k)H^T(k+1)(H(k+1)\alpha(k+1 \mid k)H^T(k+1) + R(k+1))^{-1}H(k+1)\alpha(k+1 \mid k)$$

### *Data Processing for LDS with Reducible PU (Identification Problem)*

Let continue above consideration. Let suppose that $H(k+1)$ matrix is of following block structure $H(k+1) = [H_1(k+1), \ 0]$, i.e. part of $x(k+1)$ components are not measured directly. Let divide $x(k+1)$ vector into two parts in such fashion $z(k+1) = H(k+1)x(k+1) = H_1(k+1)x_1(k+1)$. Corresponding sharing of $\bar{x}(k+1)$ and $\alpha(k+1)$ are $\bar{x}(k+1) = \begin{bmatrix} \bar{x}_1(k+1) \\ \bar{x}_2(k+1) \end{bmatrix}$ and $\alpha(k+1) = \begin{bmatrix} \alpha_{11}(k+1) & \alpha_{21}(k+1) \\ \alpha_{12}(k+1) & \alpha_{22}(k+1) \end{bmatrix}$.

Then, in accordance with [1], the following must be true

$$\bar{x}_1(k+1 \mid k+1) = \bar{x}_1(k+1) + \alpha_{11}(k+1)H_1^T(k+1)[H_1(k+1)\alpha_{11}(k+1)H_1^T(k+1) + R(k+1)]^{-1}(z(k+1) -$$
$$- H_1(k+1)\bar{x}_1(k+1))$$
$$\bar{x}_2(k+1 \mid k+1) = \bar{x}_2(k+1) + \alpha_{21}(k+1)H_1^T(k+1)[H_1(k+1)\alpha_{11}(k+1)H_1^T(k+1) + R(k+1)]^{-1}(z(k+1) -$$
$$- H_1(k+1)\bar{x}_1(k+1))$$
$$\alpha_{11}(k+1 \mid k+1) = \alpha_{11}(k+1) -$$
$$- \alpha_{11}(k+1)H_1^T(k+1)[H_1(k+1)\alpha_{11}(k+1)H_1^T(k+1) + R(k+1)]^{-1}H_1(k+1)\alpha_{11}(k+1)$$

$$\alpha_{21}(k+1|k+1) = \alpha_{21}(k+1) -$$
$$- \alpha_{21}(k+1)H_1^T(k+1)[H_1(k+1)\alpha_{11}(k+1)H_1^T(k+1) + R(k+1)]^{-1}H_1(k+1)\alpha_{11}(k+1)$$
$$\alpha_{22}(k+1|k+1) = \alpha_{22}(k+1) -$$
$$- \alpha_{21}(k+1)H_1^T(k+1)[H_1(k+1)\alpha_{11}(k+1)H_1^T(k+1) + R(k+1)]^{-1}H_1(k+1)\alpha_{12}(k+1)$$

From this it is inferred that non-measured components of $x(k+1)$ will be identified via covariance with measured elements. If assume that for $\{x(k+1),F(k+1),G(k+1),z(k+1)\}$ set analogous relations are correct, then the following should be true

$$\overline{x}(k+1|k+1) = \overline{x}(k+1|k) + \alpha(k+1|k)B(k+1)$$
$$\overline{F_j^i}(k+1|k+1) = \overline{F_j^i}(k+1|k) + \varphi_{jk}^i(k+1|k)B^k(k+1)$$
$$\overline{G_j^i}(k+1|k+1) = \overline{G_j^i}(k+1|k) + \psi_{jk}^i(k+1|k)B^k(k+1))$$
$$\alpha(k+1|k+1) = \alpha(k+1|k) - \alpha(k+1|k)A(k+1)\alpha(k+1|k)$$
$$\beta_{jl}^{ik}(k+1|k+1) = \beta_{jl}^{ik}(k+1|k) - \varphi_{jm}^i(k+1|k)A_n^m(k+1)\overline{\varphi}_l^{nk}(k+1|k)$$
$$\gamma_{jl}^{ik}(k+1|k+1) = \gamma_{jl}^{ik}(k+1|k) - \psi_{jm}^i(k+1|k)A_n^m(k+1)\overline{\psi}_l^{nk}(k+1|k)$$
$$v_{jl}^{ik}(k+1|k+1) = v_{jl}^{ik}(k+1|k) - \varphi_{jm}^i(k+1|k)A_n^m(k+1)\overline{\psi}_l^{nk}(k+1|k)$$
$$\varphi_{jk}^i(k+1|k+1) = \varphi_{jk}^i(k+1|k) - \varphi_{jm}^i(k+1|k)A_n^m(k+1)\{\alpha^T(k+1|k)\}_k^n$$
$$\psi_{jk}^i(k+1|k+1) = \psi_{jk}^i(k+1|k) - \psi_{jm}^i(k+1|k)A_n^m(k+1)\{\alpha^T(k+1|k)\}_k^n$$

here

$$A(k+1) = H^T(k+1)(H(k+1)\alpha(k+1|k)H^T(k+1) + R(k+1))^{-1}H(k+1)$$
$$B(k+1) = H^T(k+1)(H(k+1)\alpha(k+1|k)H^T(k+1) + R(k+1))^{-1}(z(k+1) - H(k+1)\overline{x}(k+1|k))$$

It should be noted that above data processing methods for LDS with PUs allow estimate the adequacy measure of applied mathematical model and treated object itself

$$\theta(k+1) = (z(k+1) - \overline{z}(k+1|k))^T (R(k+1) + H(k+1)\alpha(k+1|k)H^T(k+1))^{-1}(z(k+1) - \overline{z}(k+1|k))$$

where $\overline{z}(k+1|k) = H(k+1)\overline{x}(k+1|k)$. Value of $\theta(k)$ is distributed in accordance with $\chi^2$ law with $L$ degrees of freedom [1], and $\sqrt{2\theta^2(k)} - \sqrt{2L-1}$ value roughly obeys to normal law. If there in the $\theta(k)$ sequence of estimates frequently appear the samples of high amplitude, this is a sign that probably model is not corresponding to studied object now.

When a new-created system is experimentally studied there often appears a need to obtain some a priori estimates, which will characterize effectiveness of applied testing modes. Let suppose that experimental study of LDS with reducible PU implies application to object under testing some sequence of controls $U_0^{k-1} = \{u(0),...,u(k-1)\}$ with a purpose to improve $\{F,G\}$ estimates of inexact known parameters. One of possible criteria of test effectiveness may be $I_{FG}(k)$ predicted information content about $\{F,G\}$, which may be extracted from measurements when will be solved the identification problem.

Information content, which should contain $Z_1^k = \{z(1),...,z(k)\}$ measurements, is defined as

$$I_{FG}(Z_1^k) = S_{FG}(0) - S_{FG}(Z_1^k) \tag{12}$$

here $S_{FG}(0)$ - a priory and $S_{FG}(Z_1^k)$ - a posteriori entropy of $\{F,G\}$ estimates. A posteriori estimates must be computed after application to researched object $U_0^{k-1}$ sequence and follow-up processing of $Z_1^k$ measurements set.

It may be written

$$I_{FG}(Z_1^k) = S_{FG}(0) - S_{FG}(Z_1^{k-1}) + S_{FG}(Z_1^{k=1}) - S_{FG}(Z_1^k) = I_{FG}(Z_1^{k-1}) + I_{FG}(z(k)|Z_1^{k-1}) \tag{13}$$

here $I(z(k)|Z_1^{k-1}) = S_{FG}(Z_1^{k-1}) - S_{FG}(Z_1^k)$ - information content that is presented in $z(k)$ measurement under condition that $Z_1^{k-1}$ measurements sequence is received and processed. (13) represents a recurrent relation for computation $I_{FG}(Z_1^k)$ estimate; initial value will be $I_{FG}(Z_1^1) = S_{FG}(0) - S_{FG}(z(1))$.

Since considered problem is to determine $I_{FG}(k)$ predicted estimate (this notation underline the fact that estimates must be calculated without receiving and processing real measurements), above relation may be rewritten as

$$I_{FG}(k) = I_{FG}(k-1) + I_{FG}(z(k)|I_{FG}(k-1)) \tag{14}$$

where $I(z(k)|I_{FG}(k-1))$ - predicted information content which should be contained in $z(k)$ under condition that by the time $k$ it is predicted to accumulate $I_{FG}(k-1)$ information; initial value for this relation is $I_{FG}(1) = I_{FG}(z(1)) = S_{FG}(0) - S_{FG}(z(1))$ as well.

Let determine $I_{FG}(z(1))$ estimate. On a base of $I_A(B) = I_B(A)$ property of information content, it may be written $I_{FG}(z(1)) = I_{z(1)}(F,G)$, where $I_{z(1)}(F,G) = S(z(1)) - S(z(1)|F,G)$. Since model of system assumes that all estimates are distributed in accordance with normal law, the following must be true

$$S(z(1)) = \ln\det(\mathrm{var}(z(1))) = \ln\det(H(1)\alpha(1)H^T(1) + R(1)) \tag{15}$$
$$S(z(1)|F,G) = \ln\det(\mathrm{var}(z(1)|F,G)) = \ln\det(H(1)\alpha(1|F,G)H^T(1) + R(1)) \tag{16}$$

here: $\alpha(1)=\alpha(1|0)$ is predicted covariance matrix for system with inexact known parameters, and $\alpha(1|F,G)$ is predicted covariance matrix for system states with $F(0) = \overline{F}(0)$ and $G(0) = \overline{G}(0)$ exact parameters, i.e. prediction of Kalman filter. Predictions (15) and (16) must be calculated with respect to $P_R(0) = \{\overline{x}(0), \overline{F}(0), \overline{G}(0), \alpha(0), \beta(0), \gamma(0), \nu(0), \varphi(0), \psi(0)\}$ information state.

Let consider $A$ and $\Delta A$ square matrix of ($NxN$) dimensions, at that $A$ и $A+\Delta A$ matrixes are non-singular and $\|A\|>>\|\Delta A\|$. Then will be true $\ln\det(A+\Delta A) \cong \ln\det A + \frac{1}{N}SpA^{-1}\Delta A$ approximation. Using this approximation it is possible to derive the following relation for one-step prediction of information content

$$I_{FG}(z(1)) = I_{z(1)}(F,G) \approx \frac{1}{L}Sp(H(1)\alpha(1|F,G)H^T(1) + R(1))^{-1}H(1)\Delta\alpha(1)H^T(1) \qquad (17)$$

here: $L$ – dimension of measurements vector, $\Delta\alpha(1) = \alpha(1) - \alpha(1|F,G)$ - variation of $\alpha(1)$ prediction with respect to nominal value, which is given by $\alpha(1|F,G)$ matrix. If assume that $\varphi(0) = 0$ and $\psi(0) = 0$ then

$$\Delta\alpha(1) = \left\{\beta_{lj}^{ik}(0)(\alpha_k^l(0) + \overline{x}^l(0)\overline{x}_k(0)) + \gamma_{lj}^{ik}(0)u^l(0)u_k(0)\right\} + A + A^T \qquad (18)$$

where $A = \left\{v_{lj}^{ik}(0)\overline{x}^l(0)u_k(0)\right\}$

Relations (17) and (18) determine the estimate of predicted information content about inexact known parameters, which will be contained in $z(1)$ measurement. Derived relations may be applied for calculation of multistep prediction of type $I_{FG}(z(k)) = I_{FG}(z(k)|I_{FG}(k-1)=0)$ as well.

In general for computation of such predictions there must be applied formulas for LDS with reducible PU; however here may be accepted following PU model – up to $k$-1 point of time PU of system is considered as irreducible, $\varphi(.)$ and $\psi(.)$ items of information state are ignored, and on a base of given $U_0^{k-1}$ trajectory the $P_I(i)$, $i=1...k$-1 multistep predictions are computed. Then, starting from this point of time, PU is considered as reducible, it is assigned $\varphi(k-1) = 0$ and $\psi(k-1) = 0$, and to be computed $I_{FG}(z(k))$ estimate.

However above procedure have one essential disadvantage – it may be shown that caeteris paribus the inequalities $\|\alpha_I(k|P_I(0))\| > \|\alpha_R(k|P_R(0))\|$ and $\|\alpha_I(k|P_I(0))\| > \|\alpha_R(k|Z_1^{k-1},P_R(0))\|$ are true. From this follows that resulted estimate will be essentially reduced.

It is possible to apply another model for calculation the prediction of information content. This model assumes that up to $k$-1 point of time there is no PUs, and parameters values equal to a priory expectations, i.e. $F(0) = \overline{F}(0)$ and $G(0) = \overline{G}(0)$. In this case reasonably to use Kalman filter and, as covariance matrix of Kalman filter is independent of measurements, it is possible to accept that $\alpha(k|P_I(0)) = \alpha_{KF}(k|Z_1^{k-1},P_R(0))$. Then it is supposed that at $k$-1 point of time parameters take on some random values, which statistical characteristics are determined by corresponding items of $P_I(0)$ (at that assuming $\varphi(k-1) = 0$ and $\psi(k-1) = 0$), and $I_{FG}(z(k))$ estimate is calculated. This procedure will provide increased values of prediction estimate, i.e. $\|\alpha_{KF}(k|Z_1^{k-1}P_I(0))\| < \|\alpha_R(k|Z_1^{k-1},P_R(0))\|$.

Estimate, which will be computed in accordance with first procedure, will be referred to as "pessimistic", and this one will be used for evaluation lower edge of $I_{FG}(z(k))$. Second procedure will provide "optimistic" estimate that will be applied for estimation of $I_{FG}(z(k))$ upper edge. Both these estimates are given by

$$I_{FG}(z(k)) = I_{z(k)}(F,G) = \frac{1}{L}Sp(H(k)\alpha(k|F,G)H^T(k) + R(k))^{-1}H(k)\Delta\alpha(k)H^T(k) \qquad (19)$$

$$\Delta\alpha(k) = \left\{\beta_{lj}^{ik}(0)(\alpha_k^l(k-1|0) + \bar{x}^l(k-1|0)\bar{x}_k(k-1|0)) + \gamma_{lj}^{ik}(0)u^l(k-1)u_k(k-1)\right\} \quad (20)$$
$$+ A + A^T$$

here $A = \left\{v_{lj}^{ik}(0)\bar{x}^l(k-1|0)u_k(k-1)\right\}$, $\alpha(k|F,G)=\alpha(k|k-1,F,G)$ – Kalman filter prediction which is calculated in accordance with adopted model of parameters behavior ("pessimistic" or "optimistic" estimates). Both models compute $\bar{x}(k-1|0)$ estimate according to multistep prediction for LDS with irreducible PU.

Model of system implies that parameters estimates are normally distributed, therefore $S_{FG}(.)$ entropy, by analogy to well-known definition of multidimensional normal distribution will look like $S_{FG}(.) = \ln(\det(\Xi(.)))$, $\Xi(.) = \{\beta(.), \gamma(.), v(.)\}$. Treatment of concrete form of determinant function for the such data set is out of this paper scope, however for providing further consideration it is necessary to accept that $\det(\Xi(.))$ function possess all properties as standard determinant of matrix analyses. Then it is possible to write

$$S_{FG}(k) = \ln\det(\Xi(k)) = \ln\det(\Xi(0)) - I_{FG}(k) = \ln(\exp(-I_{FG}(k))\det(\Xi(0))) +$$
$$= \ln\det(\exp(-I_{FG}(k)/P)\Xi(0))$$

here $P$ – number of inexact known parameters of $\{F,G\}$ set. If suppose that from equation $\det(\Xi(k)) = \det(\exp(-I_{FG}(k)/P)\Xi(0))$ follows $\Xi(k) = \exp(-I_{FG}(k)/P)\Xi(0)$ then, replacing $\Xi(0)$ item with $\Xi(k-1) = \exp(-I_{FG}(k-1)/P)\Xi(0)$ in relation for $I_{FG}(z(k))$, and taking into account the linearity property of $I_{FG}(z(k))$ over $\Xi(k)$ items, it is possible to derive

$$I_{FG}(z(k) | I_{FG}(k-1)) = \exp(-I_{FG}(k-1)/P)I_{FG}(z(k))$$

Then, taking into consideration all stated above, the following may be written

$$I_{FG}(k) = I_{FG}(k-1) + \exp(-I_{FG}(k-1)/P)I_{FG}(z(k)) \quad (21)$$

Last relation, along with (19)-(20), determines the looked for estimate for prediction of information content that should be obtained with solving identification problem along $U_0^{k-1}$ trajectory.

## Optimal Stochastic Control for LDS with PUs

Let assume that performance measure (controls cost) for LDS is described by quadratic functional

$$J(U_0^{N-1}) = \frac{1}{2} M \left\{ \sum_{k=0}^{N-1} \left( x(k+1)^T A(k+1)x(k+1) + u^T(k)B(k)u(k) \right) \right\} \qquad (22)$$

where $A(.)$ is positively semidefinite and $B(.)$ is positively definite matrixes. Problem objective is to determine $U_0^{N-1}$ controls that will minimize the (23) subject to (1)-(2) dynamic constrains.

For solving this problem hereafter will be used a procedure, which is some kind of invariant embedding [13]. This approach uses consideration of some other problem, usually superior one. As this takes place, the searched solutions will be embedded in ones of this general-type problem, which in many cases may be solved with relative ease.

### *Optimal Stochastic Control for LDS with Irreducible PU*

Optimal controls will be searched with application of stochastic dynamic programming procedure [9]-[12], which for considered case may be rewritten as

$$J^*(k) = \min_{u(k)} M \{ \frac{1}{2} (x(k+1)^T A(k+1)x(k+1) + u^T(k)B(k)u(k)) + J^*(k+1) \mid P_I(k) \} \qquad (23)$$

where solving begins from consideration of following minimization problem

$$J^*(N-1) = \min_{u(N-1)} M \{ \frac{1}{2} (x(N)^T A(N)x(N) + u^T(N-1)B(N-1)u(N-1)) \mid P_I(N-1) \} \qquad (24)$$

In both these relations $P_I(.)$ is introduced defined above information state of LDS with irreducible PU. It should be remembered that $\{\overline{F}(k), \overline{G}(k), \beta(k), \gamma(k), \nu(k)\}$, $k = 0,\ldots,N-1$ trajectories are considered as a priory exactly known.

Before proceeding further it is necessary to carry out following additional consideration.

Let consider $\alpha^*(k \mid k) = \alpha_{KF}(k \mid Z_1^{k-1}, P_I(0))$ trajectory of dispersion matrix of LDS with exactly known parameters (i.e. Kalman filter), moreover $F(k) = \overline{F}(k)$ and $G(k) = \overline{G}(k)$, $k=0,\ldots,N-1$. For considered case $\alpha^*(k \mid k)$ trajectory, as well as $\alpha^*(k+1 \mid k)$ prediction trajectory, will determine nominal ones with respect to $\alpha(k \mid k) = \alpha_I(k \mid Z_1^{k-1}, U_0^{k-1}, P_I(0))$ trajectory. Then $\Delta \alpha(k \mid k) = \alpha(k \mid k) - \alpha^*(k \mid k)$ may be defined as variation of $\alpha(k \mid k)$ with respect to $\alpha^*(k \mid k)$. Variation of $\alpha(k+1 \mid k) = \alpha(k+1 \mid P_I(k))$ prediction with respect to nominal trajectory is given by

$$\Delta \alpha(k+1 \mid k) = \alpha(k+1 \mid k) - \alpha^*(k+1 \mid k) = \overline{F}(k) \Delta \alpha(k \mid k) \overline{F}^T(k) + A + A^T +$$
$$+ \left\{ \beta_{lj}^{ik}(k)(\alpha_k^l(k \mid k) + \overline{x}^l(k \mid k) \overline{x}_k(k \mid k)) + \gamma_{lj}^{ik}(k) u^l(k) u_k(k) \right\}$$

here $A = \{v_{lj}^{ik}(k)\overline{x}^{-l}(k|k)u_k(k)\}$.

Correction of Kalman filter is given by

$$\alpha(k+1|k+1) = (\alpha^{-1}(k+1|k) + H^T(k+1)R^{-1}(k+1)H(k+1))^{-1}.$$

From this results following approximation for $\Delta\alpha(k+1|k+1)$ variation

$$\Delta\alpha(k+1|k+1) \cong D(k+1)\Delta\alpha(k+1|k)D^T(k+1) \tag{25}$$

where

$$D(k+1) = \alpha^*(k+1|k+1)(\alpha^*(k+1|k))^{-1} = (I + \alpha^*(k+1|k)H^T(k+1)R^{-1}(k+1)H(k+1))^{-1} =$$
$$= I - \alpha^*(k+1|k)H^T(k+1)(R(k+1) + H(k+1)\alpha^*(k+1|k)H^T(k+1))^{-1}H(k+1)$$

Now it is possible to come back to consideration of main problem of this section – determine the optimal controls for LDS with irreducible PU.

Let consider the following function

$$V(N-1) = \frac{1}{2}\{M\{(x(N)^T A(N|N)x(N) + u^T(N-1)B(N-1)u(N-1))|P_I(N-1)\} +$$
$$+ SpK(N|N)\Delta\alpha(N|N-1) + J_0^*(N)$$

where $K(N|N)$ is symmetrical positively semidefinite matrix and $J_0^*(N)$ is some constant. It is easy to see that under conditions $A(N|N) = A(N)$, $K(N|N) = 0$ and $J_0^*(N) = 0$ expression for $V(N-1)$ coincides with (24). Therefore under these conditions minimization of $V(N-1)$ will provide the same solution as (24).

After computation the expectation and grouping the items of obtained equation it is possible to derive the following

$$V(N-1) = \frac{1}{2}\{(x(N|N-1)^T A(N|N)x(N|N-1) + SpA(N|N)\alpha(N|N-1) +$$
$$u^T(N-1)B(N-1)u(N-1)) + SpK(N|N)\Delta\alpha(N|N-1)\} + J_0^*(N) =$$
$$= \frac{1}{2}\{(\overline{F}(N-1)\overline{x}(N-1|N-1) + \overline{G}(N-1)u(N-1))^T A(N|N)(\overline{F}(N-1)\overline{x}(N-1|N-1) + \overline{G}(N-1)u(N-1)) +$$
$$+ Sp(A(N|N) + K(N|N))\{\beta_{lj}^{ik}(N-1)(\alpha_k^l(N-1|N-1) + \overline{x}^l(N-1|N-1)\overline{x}_k(N-1|N-1)) +$$
$$+ \gamma_{lj}^{ik}(N-1)u^l(N-1)u_k(N-1) + v_{lj}^{ik}(N-1)\overline{x}^l(N-1|N-1)u_k(N-1) +$$
$$+ \overline{v}_{lj}^{ik}(N-1)u^l(N-1)\overline{x}_k(N-1|N-1)\} + SpA(N|N)(\overline{F}(N-1)\alpha(N-1|N-1)\overline{F}^T(N-1) + Q(N-1)) +$$
$$SpK(N|N)\overline{F}(N-1)\Delta\alpha(N-1|N-1)\overline{F}^T(N-1) + u^T(N-1)B(N-1)u(N-1)\} + J_0^*(N)$$

If differentiate above expression and equating with zero, it is simply to derive

$$\tilde{u}(N-1) = -C(N-1)\overline{x}(N-1|N-1)$$

where

$$C(N-1) = T^{-1}(N-1)\tilde{C}(N-1)$$
$$T(N-1) = \overline{G}^T(N-1)A(N|N)\overline{G}(N-1) + B(N-1) + \{\gamma_{jl}^{ki}(N-1)(A_k^l(N|N) + K_k^l(N|N))\}$$
$$\tilde{C}(N-1) = \overline{G}^T(N-1)A(N|N)\overline{F}(N-1) + \{\nu_{jl}^{ki}(N-1)(A_k^l(N|N) + K_k^l(N|N))\}$$

It may be deduced following equation for $V(N-1)$ by substituting $\tilde{u}(N-1)$ value

$$\tilde{V}(N-1) = \frac{1}{2}\{\bar{x}^T(N-1|N-1)A(N-1|N)\bar{x}(N-1|N-1) + Sp\tilde{K}(N-1|N)\Delta\alpha(N-1|N-1) +$$
$$+ SpL(N-1|N)\alpha(N-1|N-1) + SpA(N|N)Q(N-1)\} + J_0^*(N)$$

here

$$A(N-1|N) = \overline{F}^T(N-1)A(N|N)\overline{F}(N-1) + \{\beta_{jl}^{ki}(N-1)(A_k^l(N|N) + K_k^l(N|N))\} -$$
$$- C^T(N-1)T(N-1)C(N-1)$$
$$\tilde{K}(N-1|N) = \overline{F}^T(N-1)K(N|N)\overline{F}(N-1)$$
$$L(N-1|N) = \overline{F}^T(N-1)A(N|N)\overline{F}(N-1) + \{\beta_{jl}^{ki}(N-1)(A_k^l(N|N) + K_k^l(N|N))\}$$

Using approximation $\alpha(N-1|N-1) \cong \alpha^*(N-1|N-1) + \Delta\alpha(N-1|N-1)$ this expression may be transformed into

$$\tilde{V}(N-1) \cong \frac{1}{2}\{M(x(N-1)A(N-1|N)x(N-1)|P_I(N-1)) +$$
$$+ SpK(N-1|N)\Delta\alpha(N-1|N-1)\} + J_0^*(N-1)$$

where

$$K(N-1|N) = \overline{F}^T(N-1)K(N|N)\overline{F}(N-1) + C^T(N-1)T(N-1)C(N-1)$$
$$J_0^*(N-1) = J_0^*(N) + \frac{1}{2}\{Sp(A(N|N)Q(N-1) + C^T(N-1)T(N-1)C(N-1)\alpha^*(N-1|N-1)\}$$

If substitute $\tilde{V}(N-1)$ expression (which in essence is equation for $J^*(N+1)$) in (23) for N-2 point of time with taking into account $M(.|P_I(k-1)|P_I(k-2)) = M(.|P_I(k-2))$ relation and (25), which describes changing of $\Delta\alpha(N-1|N-1)$ variation during transfer from $P_I(N-1)$ information state to $P_I(N-2)$ one, it is possible to derive

$$J^*(N-2) = \min_{u(N-2)}\{\frac{1}{2}\{M\{(x(N-1)^T A(N-1|N-1)x(N-1) +$$
$$u^T(N-2)B(N-2)u(N-2))|P_I(N-2)\} + SpK(N-1|N-1)\Delta\alpha(N-1|N-2)\} + J_0^*(N-1)\}$$

where

$$A(N-1|N-1) = A(N-1|N) + A(N-1)$$
$$K(N-1|N-1) = D^T(N-1)K(N-1|N)D(N-1)$$

If compare the expression, which to be minimized, with one for $V(N-1)$, it is simply to see these ones are identical, at that $N$ is replaced with $V(N-1)$, $V(N-1)$ is replaced with $V(N-2)$, and for minimizing of this functional should be repeatedly applied above procedure.

Hence it is derived that optimal controls for LDS with irreducible PU are given by

$$\tilde{u}(k) = -C(k)\bar{x}(k\mid k)$$

where

$$C(k) = T^{-1}(k)\tilde{C}(k)$$
$$T(k) = \overline{G}^T(k)A(k+1\mid k+1)\overline{G}(k) + B(k) + \{\gamma_{jl}^{ki}(k)(A_k^l(k+1\mid k+1) + K_k^l(k+1\mid k+1))\}$$
$$\tilde{C}(k) = \overline{G}^T(k)A(k+1\mid k+1)\overline{F}(k) + \{v_{jl}^{ki}(k)(A_k^l(k+1\mid k+1) + K_k^l(k+1\mid k+1))\}$$

At that functional is of following reproduction form

$$V(k) = \frac{1}{2}\{M\{(x(k+1)^T A(k+1\mid k+1)x(k+1) + u^T(k)B(k)u(k))\mid P_l(k)\} +$$
$$+ SpK(k+1\mid k+1)\Delta\alpha(k+1\mid k)\} + J_0^*(k+1)$$

$A(k|k)$ и $K(k|k)$ values are determined by following recurrent relations

$$A(k\mid k) = \overline{F}^T(k)A(k+1\mid k+1)\overline{F}(k) + \{\beta_{jl}^{ki}(k)(A_k^l(k+1\mid k+1) + K_k^l(k+1\mid k+1))\} -$$
$$- C^T(k)T(k)C(k) + A(k); \quad A(N\mid N) = A(N)$$
$$K(k\mid k+1) = \overline{F}^T(k)K(k+1\mid k+1)\overline{F}(k) + C^T(k)T(k)C(k)$$
$$K(k\mid k) = D^T(k)K(k\mid k+1)D(k); \quad K(N\mid N) = 0$$

Average cost of this control is given by

$$\bar{J}^* = \frac{1}{2}\left[\bar{x}^T(0)A(0\mid 1)\bar{x}(0) + SpA(0\mid 0)\alpha(0) + \sum_{i=0}^{N-1}\{SpA(i+1\mid i+1)Q(i) + C^T(i)T(i)C(i)\alpha^*(i\mid i)\}\right] \quad (26)$$

It is easy to see that obtained solution looks like to well-known relations for CE control. When PU degree is reduced (i.e. $\Xi(k)=\{\beta(k),\gamma(k),v(k)\}\to 0$), equations for optimal controls approach to analogous relations for CE control. The most outstanding distinction lies in necessity of joint recurrent solving of two equations of Riccaty type. Appearance of this additional equation explicitly describes "precaution" property of proposed control. This may be demonstrated by following consideration.

CE controls are given by

$$C(k) = T^{-1}(k)\tilde{C}(k)$$
$$T(k) = G^T(k)A(k+1\mid k+1)G(k) + B(k)$$
$$\tilde{C}(k) = G^T(k)A(k+1\mid k+1)F(k)$$

If try to synthesize controls on a base of these relations with application of definitions that was introduced at beginning of this paper, then may be derived the following

$$\overline{T}(k) = M\{T(k)\} = \overline{G}^T(k)A(k+1|k+1)\overline{G}(k) + B(k) + \{\gamma_{jl}^{ki}(k)A_k^l(k+1|k+1)\}$$

$$\overline{\widetilde{C}}(k) = M\{\widetilde{C}(k)\} = \overline{G}^T(k)A(k+1|k+1)\overline{F}(k) + \{v_{jl}^{ki}(k)A_k^l(k+1|k+1)\}$$

i.e. in these relations disappear the items, which are proportional to $K(k+1|k+1)$. Since it may be shown that $K(k|k)$ is positively semidefinite, it is obviously that $\|T(k)\| > \|\overline{T}(k)\|$ and $\|\widetilde{C}(k)\| > \|\overline{\widetilde{C}}(k)\|$ that is equivalent to increasing of $A(k+1|k+1)$ matrix and in turns $A(k)$.

## Optimal Control for LDS with Reducible PU

Immediate deriving the relations for optimal control for LDS with reducible PU is intractable problem, what is noted by numerous authors [6,8,9]. This section contains description of approach to obtaining some approximation for optimal control that will have all properties of dual control [8,9]. Equations for optimal control will be derived on a base of above solution for LDS with irreducible PU and application of proposed model for presentation of reducible PU.

Let suppose that there is obtained the estimates of information content about inexact known parameters $I_\Sigma(k)$, $k=1,\ldots,N-1$, which will be gained via solving the identification problem along $U_0^{*N-1}$ trajectory of optimal controls. In this case may be adopted the following PU model – uncertainty is considered as irreducible, and time dependence of $\Xi(k) = \{\beta(k),\gamma(k),v(k)\}$ (in accordance with prediction of information content) is given by

$$\begin{aligned}\beta(k) &= \exp(-I_\Sigma(k)/P)\beta(0) \\ \gamma(k) &= \exp(-I_\Sigma(k)/P)\gamma(0) \\ v(k) &= \exp(-I_\Sigma(k)/P)v(0)\end{aligned} \qquad (27)$$

where $I_\Sigma(k) = \sum_{i=1}^{k} I_{FG}(i)$ and $I_{FG}(i)$ – prediction of information content that may be obtained by processing $z(k)$ measurement. Informational state for such system may be defined as $P_R^*(k) = \{\overline{x}(k|k), \alpha(k|k), I_\Sigma(k)\}$. Hereafter will be assumed that $\overline{F}(k) = \overline{F}(0), \overline{G}(k) = \overline{G}(0)$, $k=0,\ldots,N-1$.

Equation of stochastic dynamic programming is given by

$$J^*(k) = \min_{u(k)} M\{\frac{1}{2}(x(k+1)^T A(k+1)x(k+1) + u^T(k)B(k)u(k)) + J^*(k+1) | P_R^*(k)\} \qquad (28)$$

where solving is started from consideration of following minimization problem

$$J^*(N-1) = \min_{u(N-1)} M\{\frac{1}{2}(x(N)^T A(N)x(N) + u^T(N-1)B(N-1)u(N-1)) | P_R^*(N-1)\} \qquad (29)$$

Let consider the function

$$V(N-1) = \frac{1}{2}\{M\{(x(N)^T A(N|N)x(N) + u^T(N-1)B(N-1)u(N-1)) | P_R^*(N-1)\} +$$
$$+ SpK(N|N)\Delta\alpha(N|N-1)\} + J_0^*(N) + \lambda(N)I_\Sigma(N)$$

where $K(N|N)$ is symmetric positively semidefinite matrix, $\Delta\alpha(N|N-1)$ is variation of $\alpha(N|N-1)$ with respect to nominal trajectory ($\alpha^*(k+1|k)$ nominal trajectory is defined similar to one of above section), $\lambda(N)$ and $J_0^*(N)$ are some constants.

It should be reminded that in accordance with stated above, information content of $z(k+1)$ measurement is given by

$$I_{FG}(k+1) = I_{FG}(z(k+1)) \cong \frac{1}{L} Sp(H(k+1)\alpha^*(k+1|k)H^T(k+1) + R(k+1))^{-1} H(k+1)\Delta\alpha(k+1)H^T(k+1)$$

where $L$ – dimension of $R_z^L$ space,

$$\Delta\alpha(k+1) = \{\beta_{lj}^{ik}(k)(\alpha_k^l(k|k) + \bar{x}^l(k)\bar{x}_k(k)) + \gamma_{lj}^{ik}(k)u^l(k)u_k(k) + v_{lj}^{ik}(k)\bar{x}^l(k)u_k(k) + \bar{v}_{lj}^{ik}(k)u^l(k)\bar{x}_k(k)\}$$

and $\beta(k)$, $\gamma(k)$ и $v(k)$ are determined in accordance with (27).

It is simply to see that under conditions $A(N|N)=A(N)$, $K(N|N)=0$, $J_0^*(N) = 0$ and $\lambda(N)=0$ expression for $V(N-1)$ exact coincides with one for cost (29). Therefore under such conditions minimization of $V(N-1)$ will provide the same solutions as (29).

It is possible to obtain

$$V(N-1) = \frac{1}{2}\{(x(N|N-1)^T A(N|N)x(N|N-1) + SpA(N|N)\alpha(N|N-1) +$$
$$u^T(N-1)B(N-1)u(N-1)) + SpK(N|N)\Delta\alpha(N|N-1)\} + \lambda(N)I(N) + J_0^*(N) + \lambda(N)I_\Sigma(N) =$$
$$= \frac{1}{2}\{(\bar{F}(0)\bar{x}(N-1|N-1) + \bar{G}(0)u(N-1))^T A(N|N)(\bar{F}(0)\bar{x}(N-1|N-1) + \bar{G}(0)u(N-1)) +$$
$$+ Sp(A(N|N) + K(N|N) + \frac{\lambda(N)}{L}S(N))\{\beta_{lj}^{ik}(N-1)(\alpha_k^l(N-1|N-1) + \bar{x}^l(N-1|N-1)\bar{x}_k(N-1|N-1)) +$$
$$+ \gamma_{lj}^{ik}(N-1)u^l(N-1)u_k(N-1) + v_{lj}^{ik}(N-1)\bar{x}^l(N-1|N-1)u_k(N-1) +$$
$$+ \bar{v}_{lj}^{ik}(N-1)u^l(N-1)\bar{x}_k(N-1|N-1)\} + SpA(N|N)(\bar{F}(0)\alpha(N-1|N-1)\bar{F}^T(0) + Q(N-1)) +$$
$$+ SpK(N|N)\bar{F}(0)\Delta\alpha(N-1|N-1)\bar{F}^T(0) + u^T(N-1)B(N-1)u(N-1)\} + J_0^*(N) + \lambda(N)I_\Sigma(N-1)$$

here $S(N) = H^T(N)(H(N)\alpha^*(N|N-1)H^T(N) + R(N))^{-1} H(N)$

If above expression differentiate with respect to $u(N-1)$ and equating derived formula to zero, it is simply to obtain following expression for optimal control

$$\tilde{u}(N-1) = -C(N-1)\bar{x}(N-1|N-1)$$

where:

$$C(N-1) = T^{-1}(N-1)\tilde{C}(N-1)$$

$$T(N-1) = \bar{G}^T(0)A(N|N)\bar{G}(0) + B(N-1) + \{\gamma_{jl}^{ki}(N-1)(A_k^l(N|N) + K_k^l(N|N) + \frac{\lambda(N)}{L}S_k^l(N))\}$$

$$\tilde{C}(N-1) = \bar{G}^T(0)A(N|N)\bar{F}(0) + \{v_{jl}^{ki}(N-1)(A_k^l(N|N) + K_k^l(N|N) + \frac{\lambda(N)}{L}S_k^l(N))\}$$

If substitute $\tilde{u}(N-1)$ in expression for $V(N-1)$ and grouping corresponding item, it may be derived that

$$\tilde{V}(N-1) = \frac{1}{2}\{\bar{x}^T(N-1|N-1)A(N-1|N)\bar{x}(N-1|N-1) + SpK(N-1|N)\Delta\alpha(N-1|N-1) + $$
$$+ SpL(N-1|N)\alpha(N-1|N-1) + SpA(N|N)Q(N-1)\} + J_0^*(N) + \lambda(N)I_\Sigma(N-1)$$

where

$$A(N-1|N) = \bar{F}^T(0)A(N|N)\bar{F}(0) + \{\beta_{jl}^{ki}(N-1)(A_k^l(N|N) + K_k^l(N|N) + \frac{\lambda(N)}{L}S_k^l(N))\} - $$
$$- C^T(N-1)T(N-1)C(N-1)$$
$$K(N-1|N) = \bar{F}^T(0)K(N|N)\bar{F}(0)$$
$$L(N-1|N) = \bar{F}^T(0)A(N|N)\bar{F}(0) + \{\beta_{jl}^{ki}(N-1)(A_k^l(N|N) + K_k^l(N|N) + \frac{\lambda(N)}{L}S_k^l(N))\}$$

Using approximation $\alpha(N-1|N-1) \cong \alpha^*(N-1|N-1) + \Delta\alpha(N-1|N-1)$ by analogy with consideration of above section, expression for $\tilde{V}(N-1)$ may be transformed in

$$\tilde{V}(N-1) \cong \frac{1}{2}\{M(x(N-1)A(N-1|N)x(N-1)|P_R(N-1)) + SpK(N-1|N)\Delta\alpha(N-1|N-1)\} + \tilde{J}_0^*(N-1)$$

where

$$K(N-1|N) = \bar{F}^T(0)K(N|N)\bar{F}(0) + C^T(N-1)T(N-1)C(N-1)$$
$$\tilde{J}_0^*(N-1) = J_0^*(N) + \lambda(N)I_\Sigma(N-1) + $$
$$+ \frac{1}{2}\{Sp(A(N|N)Q(N-1) + C^T(N-1)T(N-1)C(N-1)\alpha^*(N-1|N-1)\}$$

Let subdivide expression for $\tilde{V}(N-1)$ into two parts – static one, which is independent from $\bar{x}(N-1|N-1)$, $\alpha(N-1|N-1)$ and $\Delta\alpha(N-1|N-1)$ estimates (i.e. $\tilde{J}_0^*(N-1)$), and dynamic one that is dependant from above estimates. Let suppose that $I_\Sigma(N-1) = 0$ for dynamic part, and let apply to it $P_I^*(N-1) \rightarrow P_I^*(N-2)$ transition method from above section.

Application of this procedure leads to following expression

$$\tilde{V}(N-1) \cong \frac{1}{2}\{M(x(N-1)A(N-1|N-1)x(N-1)|P_R^*(N-2)) + u^T(N-2)B(N-2)u(N-2) + $$
$$+ SpK(N-1|N-1)\Delta\alpha(N-1|N-2)\} + \tilde{J}_0^*(N-1)$$

where:

$$A(N-1|N-1) = A(N-1|N) + A(N-2)$$
$$A(N-1|N) = \bar{F}^T(0)A(N|N)\bar{F}(0) + \{\beta_{jl}^{ki}(0)(A_k^l(N|N) + K_k^l(N|N) + \bar{\lambda}(N)S_k^l(N))\} - $$
$$- C^T(N-1)T(N-1)C(N-1)$$
$$K(N-1|N) = D^T(N-1)(\bar{F}^T(0)K(N|N)\bar{F}(0) + C^T(N-1)T(N-1)C(N-1))D(N-1)$$

$$C(N-1) = T^{-1}(N-1)\widetilde{C}(N-1)$$

$$T(N-1) = \overline{G}^T(0)A(N|N)\overline{G}(0) + B(N-1) + \{\gamma_{jl}^{ki}(0)(A_k^l(N|N) + K_k^l(N|N) + \frac{\lambda(N)}{L}S(N)_k^l)\}$$

$$\widetilde{C}(N-1) = \overline{G}^T(0)A(N|N)\overline{F}(0) + \{v_{jl}^{ki}(0)(A_k^l(N|N) + K_k^l(N|N) + \frac{\lambda(N)}{L}S(N))\}$$

Static part saves dependence from $I_\Sigma(N-1)$, therefore may be applied following approximation

$$\widetilde{J}_0^*(N-1) \cong \widetilde{J}_0^*(N-1)_{I_\Sigma(N-1)=0} + \left[\frac{\partial \widetilde{J}_0^*(N-1)}{\partial I_\Sigma(N-1)}\right]_{I_\Sigma(N-1)=0} I_\Sigma(N-1)$$

It may be written

$$\widetilde{V}(N-1) \cong \frac{1}{2}\{M(x(N-1)A(N-1|N-1)x(N-1)|P_I^*(N-2)) + u^T(N-2)B(N-2)u(N-2) +$$
$$+ SpK(N-1|N-1)\Delta\alpha(N-1|N-2)\} + J_0^*(N-1) + \lambda(N-1)I_\Sigma(N-1)$$

where $J_0^*(N-1) = \widetilde{J}_0^*(N-1)_{I(N-1)=0}$ and

$$\lambda(N-1) = \frac{\partial \widetilde{J}_0^*(N-1)}{\partial I_\Sigma(N-1)}\bigg|_{I_\Sigma(N-1)=0} = \lambda(N) - \frac{1}{2P}Sp(\dot{C}^T(N-1)C(N-1) +$$
$$+ C^T(N-1)\dot{C}(N-1) - C^T(N-1)\dot{T}(N-1)C(N-1))\alpha^*(N-1|N-1)$$

here $P$ is number of inexact known parameters of system, and

$$\dot{T}(N-1) = \{\gamma_{jl}^{ki}(0)(A_k^l(N|N) + K_k^l(N|N) + \frac{\lambda(N)}{L}S_k^l(N))\}$$

$$\dot{C}(N-1) = \{v_{jl}^{ki}(0)(A_k^l(N|N) + K_k^l(N|N) + \frac{\lambda(N)}{L}S_k^l(N))\}$$

If compare expressions for $\widetilde{V}(N-1)$ and $V(N)$, it is simply to see that they are identical (except indexes); $N$ is replaced with $N$-2, $N$-1 with $N$-2, i.e. for minimizing of $\widetilde{V}(N-1)$ may be applied above procedure.

And this in one's part means that optimal control for LDS with reducible PU are given by

$$\widetilde{u}(k) = -C(k)\overline{x}(k|k)$$

where:
$$C(k) = T^{-1}(k)\widetilde{C}(k)$$
$$T(k) = \overline{G}^T(0)A(k+1|k+1)\overline{G}(0) + B(k) + \{\gamma_{jl}^{ki}(0)(A_k^l(k+1|k+1) + K_k^l(k+1|k+1) +$$
$$+ \frac{\lambda(k+1)}{L}S_k^l(k+1))\}$$
$$\widetilde{C}(k) = \overline{G}^T(0)A(k+1|k+1)\overline{F}(0) + \{v_{jl}^{ki}(0)(A_k^l(k+1|k+1) + K_k^l(k+1|k+1) +$$
$$+ \frac{\lambda(k+1)}{L}S_k^l(k+1))\}$$

$$S(k+1) = H^T(k+1)(H(k+1)\alpha^*(k+1|k)H^T(k+1) + R(k+1))^{-1} H(k+1)$$

For these controls reproduction form of functional is given by

$$V(k) = \frac{1}{2}\{M\{(x(k+1)^T A(k+1|k+1)x(k+1) + u^T(k)B(k)u(k)) | P_I(k)\} +$$
$$+ SpK(k+1|k+1)\Delta\alpha(k+1|k)\} + \lambda(k+1)I_\Sigma(k+1) + J_0^*(k+1)$$

$A(k|k)$, $K(k|k)$ and $\lambda(k)$ are determined by recurrent relation

$$A(k|k) = \overline{F}^T(0)A(k+1|k+1)\overline{F}(0) + \{\beta_{jl}^{ki}(k)(A_k^l(k+1|k+1) + K_k^l(k+1|k+1) +$$
$$+ \frac{\lambda(k+1)}{L} S_k^l(k+1))\} - C^T(k)T(k)C(k) + A(k); \quad A(N|N) = A(N)$$
$$K(k|k+1) = \overline{F}^T(0)K(k+1|k+1)\overline{F}(0) + C^T(k)T(k)C(k)$$
$$K(k|k) = D^T(k)K(k|k+1)D(k); \quad K(N|N) = 0$$
$$\lambda(k) = \lambda(k+1) - \frac{1}{2P} Sp(\dot{C}^T(i)C(i) + C^T(i)\dot{C}(N-1) - C^T(i)\dot{T}(i)C(i))\alpha^*(i|i); \quad \lambda(N) = 0$$

Average cost of application of this control law is given by (26) as well.

It should be noted that there to be applied not all $\widetilde{U}_0^{N-1} = (\widetilde{u}(0),...,\widetilde{u}(N-1))$ controls, but only $\widetilde{u}(0)$. After application $\widetilde{u}(0)$ control, there must be measured $z(1)$ input data. This measurement to be processed in accordance with proposed identification procedure, and to be obtained new $\{\overline{F}(1), \overline{G}(1)\}$ estimates, and items of informational state, which describes PU degree (i.e. $\Xi(1) = \{\beta(1), \gamma(1), \nu(1)\}$. This means that there it is necessary anew compute the nominal trajectory $\alpha^*(k|k)$, and to solve inverse recurrent relations for $A(k|k)$, $K(k|k)$ and $\lambda(k)$.

If compare the derived relations for optimal control with reducible and irreducible PU, it is possible to note that main difference lies in appearance of a new item in formulas for systems with reducible PU, which is proportional to $\lambda(k)S(k)$. It may be demonstrated that $\lambda(k)<0$ and $S(k)$ is positively semidefinite matrix. Therefore this item describes "study" effect that resides to dual control [8,9], and which should reduce influence of "precaution" factor.

It should be noted that under proposed approach there may be derived more precise, but more complicated solutions for dual control problem. If will be assumed that $A(N|N)$ and $K(N|N)$ are some function with respect to $I_\Sigma(N)$, and during transition form $P_I^*(N-1)$ to $P_I^*(N-2)$ information state these dependences should be taken into account, then $A(N-1|N-1)$ and $K(N-1|N-1)$ will depend on $I_\Sigma(N-1)$. This fact must be considered when $\lambda(N)$ to be computed, that leads to necessity compute derivates $\dot{A}(N|N) = \frac{\partial A(N|N)}{\partial I_\Sigma(N)}$ and $\dot{K}(N|N) = \frac{\partial K(N|N)}{\partial I_\Sigma(N)}$ (originally this items are considered as equal to zero). $\lambda(N-1)$ coefficient saves dependence on $\dot{A}(N-1|N-1) = \frac{\partial A(N-1|N-1)}{\partial I_\Sigma(N-1)}$ and $\dot{K}(N-1|N-1) = \frac{\partial K(N-1|N-1)}{\partial I_\Sigma(N-1)}$, which are no equal to zero as well, i.e. for computing the $\lambda(k)$ value there appears a need to solve recurrent relations for $\dot{A}(k|k)$ and $\dot{K}(k|k)$. These trajectories will be needed only for computation of $\lambda(k)$ trajectory; therefore appropriateness of such type procedure should be proved for each specific case.

## Optimal Control for Experimental Study of LDS with Reducible PU.

This section provides a common-type outline of approach to solving a new problem – optimal control for experimental study of LDS with reducible PU. Let assume that during experimental study of LDS with reducible PU there it is planned to apply as nominal testing regime some $R_1^N = \{r(1),...,r(N)\}$ trajectory of states (it should be noted that within this consideration the term experimental study implies application to researched object some controls with a purpose of improvement of estimates of $\{F,G\}$ inexact known parameters). Also let suppose that there is defined some resource that looks like

$$J(U_0^{N-1}) = \frac{1}{2} M \left\{ \sum_{k=0}^{N-1} \left( (x(k+1) - r(k+1))^T A(k+1)(x(k+1) - r(k+1)) + u^T(k)B(k)u(k) \right) \right\}$$

which must meet to restriction of type $J \leq J_0$, where $J_0$ – some predetermined constant.

Let assume that preliminary evaluation shows that if there will be applied optimal control of above section, $\bar{J}^*(R_1^N)$ value will be rather lower then $J_0$. Then it is possible to state the following problem: to determine $U_0^{N-1} = \{u(0),...,u(N-1)\}$ controls, which will provide maximal value of predicted information content about $\{F,G\}$, under $J \leq J_0$ condition.

If Lagrange multiplier technique to be applied, the functional for stated problem will be $\bar{J} = I_\Sigma + \mu (J_0 - J) \to \max$ or in differential form (hereafter $dS$ – differential of entropy of $\{F,G\}$ estimates) $d\bar{J} = dS - \mu \, dJ \geq 0$. Last relation may be rewritten as

$$TdS \geq dJ \qquad (30)$$

here $T = 1/\mu$.

If $J$ will be regarded as some generalized energy, and $T$ factor as some common-type temperature, then (30) may be considered as some kind of law of degradation of energy for problem of optimal control for experimental study of LDS with reducible PU. Simultaneously it should be taken into account the following - law of degradation energy describes physical processes that leads to increasing of entropy, but (30) describes optimal control, which should provide decreasing (more exact – minimization) of entropy of parameters estimates. It is easy to see that assigning $T = -\lambda$ the similar expression may be written for problem of previous section. At that solution of considered problem might be regarded as some generalized isotherm, and previous one – as some generalized adiabatic process.

Computed optimal controls may be applied for iterated improvement of $R_1^N$ nominal trajectory (with fixing of $A(k)$ and $B(k)$ matrixes) with a purpose of choosing of most informative test regime. When such procedure will be realized, for each anew computed $R_1^N(k)$ trajectory (here $k$ is number of iteration) would correspond anew $T(k)$ value. Trajectories, which will be determined with application of this procedure, will be characterized by $C(k) = T(k)\dfrac{dS(k)}{dT(k)}$ coefficient, which may be considered as some generalized thermal capacity from thermodynamic point of view. At that as a most effective trajectory will be regarded the regime, which will provide highest value of thermal capacity.

It is possible adduce some other thermodynamic analogies for such stochastic problems. It seems that application of well-proven thermodynamic methodology for solving the problems of stochastic optimization of LDS with PU will provide not only deepest understanding of derived solutions, but may lead to appearance of new approaches to solving of known problems and stating of anew ones.

### *Simulation Study*

Simulation was carried out in order to prove a functionality of proposed methods of data processing and optimal control for LDS with PU, as well as to obtain the some estimates of performance capabilities. As a base for this study the example of well-known simulation E. Tse and Y. Bar-Shalom ("An Actively Adaptive Control for Linear Systems with Random Parameters via Dual Control Approach", IEEE Trans. Automat. Contr., vol. AC-18, pp. 109-117, April 1973) was chosen, which have researched the third order LDS with six inexact known parameters:

$$x(k+1) = F(\theta)x(k) + G(\theta)u(k) + w(k)$$
$$z(k) = Hx(k) + v(k)$$

where

$$F(\theta) = \begin{vmatrix} 0 & 1 & 0 \\ 0 & 0 & 1 \\ \theta_1 & \theta_2 & \theta_3 \end{vmatrix}, \quad G(\theta) = \begin{vmatrix} \theta_4 \\ \theta_5 \\ \theta_6 \end{vmatrix}, \quad H = \begin{vmatrix} 0 & 0 & 1 \end{vmatrix}$$

A priory estimates of parameters were changed in comparison with Tse and Bar-Shalom simulation due to various reasons (and some cost factors as well), and were applied following a priory estimates of system parameters: $\{\theta_i\}_{i=1}^{6}$ are unknown parameters with following statistic characteristics $\bar{\theta}_i$=(1.51,-0.89,0.3,0.22,0.57,0.77) and $\sigma_\theta^2$=diag(0.1,0.1,0.1,0.01,0.01,0.1). At that true values of parameters were as follows $\theta$=(1.8,-1.01,0.58,0.3,0.5,1.0). A priory estimates of system initial states were considered as normal distributed and $\bar{x}(0) = x(0) = 0$, $\sigma_x^2 = diag\{10,10,10\}$. The noises $\{w(k)\}_{i=1}^{3}$ and $v(k)$ was assumed as independent and normally distributed with zero mean and unit dispersion.

During simulation the overall testing of both data processing and optimal control methods were carried out. Below listed the legends used for definition the applied methods:
Data processing:
- SYS = 0 - classic Kalman filter;
- SYS = 1 - filtering procedure for LDS with irreducible PU;
- SYS = 2 - procedure of estimation of states and parameters for LDS with reducible PU.

Optimal controls:
- CNT = 0 - Certainty Equivalence (CE) control;
- CNT = 1 - optimal control for LDS with irreducible PU;
- CNT = 2 - optimal control for LDS with reducible PU.

For all types of optimal controls, when was applied SYS=2 data processing procedure, after updating the estimates, the solutions of control problem were computed anew. For estimation the unattainable lower edge of cost this study includes the case of optimal control of system with exact known parameters. Simulation is represented by three tests sets each of 1000 runs of modeling program and corresponding data processing/optimal control procedures in accordance with Monte-Carlo technique (multiple modeling sets are used for evaluation of statistical stability of obtained results).

It should be noted that modern approaches to adaptive control, based on neural network and fuzzy system applies the following structure: an approximator ("identifier") that is used to estimate unknown system parameters and a "certainty equivalence" control scheme in which the system controller is defined assuming that the parameter estimates are true values. From this follows that variant SYS=2/CNT=0 may be considered as some equivalent to modern adaptive control techniques.

Interception-Type Example

In this case it is necessary to bring only the third component of state vector to a specified value. This may be expressed by the following functional (cost function)

$$J = \frac{1}{2} M \left\{ [x_3(N) - \rho]^2 + \sum_{i=0}^{N} \lambda u^2(i) \right\}$$

During simulation the following values of cost variables were applied: $N=25$, $\rho=20$, $\lambda=0.0001$. The simulation outcomes are adduced in Tab. 3.

| Variant | Cost | | | Miss distance | | |
|---|---|---|---|---|---|---|
| | Mean | Dispersion | Maximal value | Mean | Dispersion | Maximal value |
| Exact known parameters | 5,888059 | 7,484635 | 45,81559 | 11,73652 | 14,96861 | 91,59195 |
| | 5,383179 | 7,402981 | 59,77968 | 10,72696 | 14,80701 | 119,5427 |
| | 5,850405 | 8,114829 | 61,18062 | 11,66232 | 16,22958 | 122,3254 |
| SYS = 0 CNT = 0 | 1026,427 | 1448,565 | 9939,457 | 2042,248 | 2886,615 | 19807,14 |
| | 1087,65 | 1565,135 | 13051,15 | 2164,438 | 3120,213 | 26022,58 |
| | 1040,9 | 1462,184 | 10505,17 | 2070,508 | 2912,873 | 20989,19 |
| SYS = 1 CNT = 0 | 119,4912 | 155,7593 | 1147,966 | 230,0467 | 306,8111 | 2279,334 |
| | 117,7874 | 168,3559 | 1596,886 | 226,7912 | 331,228 | 3147,399 |
| | 130,0427 | 165,914 | 1209,182 | 251,0845 | 326,5649 | 2345,901 |
| SYS = 2 CNT = 0 | 28,74616 | 49,61717 | 553,5978 | 56,91493 | 99,22476 | 1106,92 |
| | 25,56459 | 37,66148 | 362,8829 | 50,55733 | 75,33169 | 724,9382 |
| | 26,54427 | 43,07641 | 507,1032 | 52,55798 | 86,12724 | 1013,301 |
| SYS = 1 CNT = 1 | 14,69206 | 18,2488 | 143,8312 | 28,88613 | 36,43652 | 284,5043 |
| | 14,20943 | 17,50404 | 127,342 | 27,93525 | 34,94536 | 253,1635 |
| | 14,99088 | 19,59168 | 149,6159 | 29,47868 | 39,11218 | 298,1324 |
| SYS = 2 CNT = 1 | 8,93055 | 12,6449 | 137,1522 | 17,50837 | 25,30189 | 273,6835 |
| | 9,450505 | 13,96352 | 148,3363 | 18,54674 | 27,94484 | 296,5401 |
| | 9,838187 | 13,90041 | 115,709 | 19,329 | 27,81682 | 230,9664 |
| SYS = 2 CNT = 2 | 17,8494 | 20,94684 | 244,0217 | 11,68999 | 15,79062 | 131,2081 |
| | 16,32223 | 22,80032 | 332,1621 | 10,22481 | 13,49113 | 114,6434 |
| | 16,70631 | 19,11658 | 204,1003 | 11,93092 | 16,53112 | 118,0493 |

*Tab. 3. Simulation outcomes for interception example*

Simulation reveals the following feature of proposed optimal control for LDS with reducible PU (SYS=2/CNT=2) – in initial phase of controls computing in 100% cases there appeared the degenerate modes, i.e. controls tended to singular ones. Due to this reason the applied control algorithm were upgraded in a following way: if computation reduced to singular modes, then were applied probe (study) controls, which were the samples from white noise with zero mean and $\sigma_u=40$ (this value were determined experimentally in accordance with criteria of minimization of resulted cost). This upgraded algorithm demonstrated rather good performances. Fig. 3 represents normalized frequency of appearance of singular controls for upgraded algorithm.

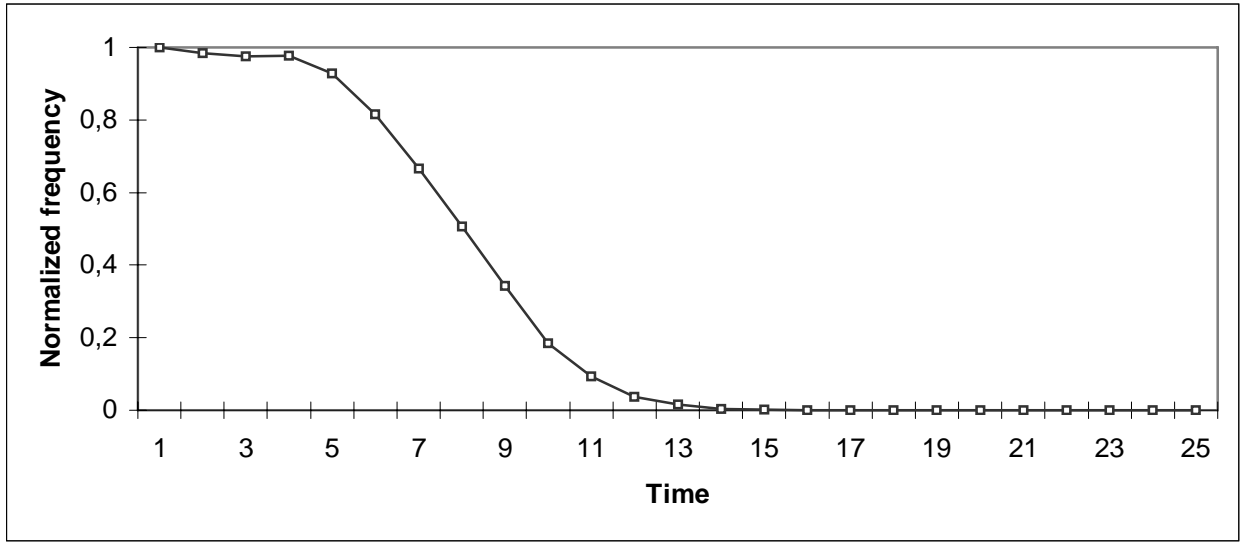

*Fig. 3. Normalized frequency of appearance the singular controls for interception example.*

Soft Landing-Type Example

In this example the problem is to bring a final state to a certain position in state space. Following expression for a cost corresponds to this problem:

$$J = \frac{1}{2} M \left\{ (x(N) - \rho)^T (x(N) - \rho) + \sum_{i=0}^{N} \lambda u^2(i) \right\}$$

where $\rho$ is some point in $R^3$, $\lambda = 0.0001$ and $N = 25$, as for previous example. This problem may be considered as a soft landing one with a final state of $\rho = \{0,0,20\}$. Simulation technique was similar to applied one for previous example. Simulation outcomes are adduced in Tab. 4.

| Variant | Cost | | | Miss distance | | |
|---|---|---|---|---|---|---|
| | Mean | Dispersion | Maximal value | Mean | Dispersion | Maximal value |
| Exact known parameters | 15,51027 | 13,38199 | 105,4067 | 30,85448 | 26,76252 | 210,5442 |
| | 14,94398 | 12,55293 | 89,83703 | 29,72488 | 25,10282 | 179,486 |
| | 15,96678 | 14,29496 | 102,9345 | 31,76713 | 28,58542 | 205,6381 |
| SYS = 0 CNT = 0 | 346,4404 | 259,1054 | 1610,78 | 691,805 | 517,4676 | 3217,353 |
| | 342,5964 | 259,7556 | 1896,751 | 684,1271 | 518,7756 | 3787,964 |
| | 346,7325 | 254,2953 | 1779,128 | 692,3976 | 507,8647 | 3553,189 |
| SYS = 1 CNT = 0 | 156,5742 | 153,5712 | 1228,505 | 312,4046 | 306,6053 | 2451,252 |
| | 170,0257 | 170,0793 | 1244,285 | 339,2673 | 339,5878 | 2484,11 |
| | 168,341 | 160,7243 | 1053,697 | 335,9233 | 320,9179 | 2103,867 |
| SYS = 2 CNT = 0 | 28,28068 | 32,14616 | 418,2113 | 56,15731 | 64,25007 | 834,2893 |
| | 26,24313 | 26,34425 | 203,956 | 52,0931 | 52,67868 | 407,6686 |
| | 27,38011 | 26,59816 | 188,1399 | 54,35755 | 53,18868 | 376,2076 |
| SYS = 1 CNT = 1 | 215,64 | 199,1483 | 1256,395 | 430,6497 | 397,8292 | 2509,844 |
| | 201,6579 | 188,013 | 1374,553 | 402,7142 | 375,5729 | 2745,214 |
| | 197,0601 | 197,3259 | 1554,117 | 393,5267 | 394,1835 | 3103,439 |
| SYS = 2 CNT = 1 | 23,54357 | 21,64384 | 188,8178 | 46,70776 | 43,296 | 377,4093 |
| | 24,95453 | 23,73557 | 189,8572 | 49,53861 | 47,47787 | 379,2981 |
| | 24,0663 | 21,03832 | 150,4909 | 47,74397 | 42,08541 | 300,6429 |
| SYS = 2 CNT = 2 | 20,58895 | 18,16012 | 230,6452 | 34,73497 | 32,00234 | 290,5604 |
| | 20,3554 | 17,60179 | 251,6055 | 34,05008 | 30,64774 | 201,0259 |
| | 20,28283 | 16,56067 | 140,6904 | 34,33234 | 31,37584 | 274,5954 |

*Tab. 4. Simulation outcomes for soft landing example*

As for case of interception-type example, when SYS=2/CNT=2 variant was computed, there appeared singular modes. As above in this case were applied probe controls, which were samples from white noise with zero mean and $\sigma_u=10$ (this value were determined experimentally in accordance with criteria of minimization of obtained cost as well). This control algorithm showed up the best performances, verge towards CE controls with exact known parameters. For this reason for clarification the processes, which take place when this control was applied, in addition to normalized frequency of appearance the singular controls, below are adduced following figures:

- Trajectories of mean and dispersion of $-\lambda(k)$ factor (if should be applied a thermodynamic analogy, this coefficient may be interpreted as some general-type temperature of studied process).
- Trajectories of mean and dispersion of current weighted control energies;
- Trajectories of mean and dispersion of real accumulated information content about inexact known parameters.

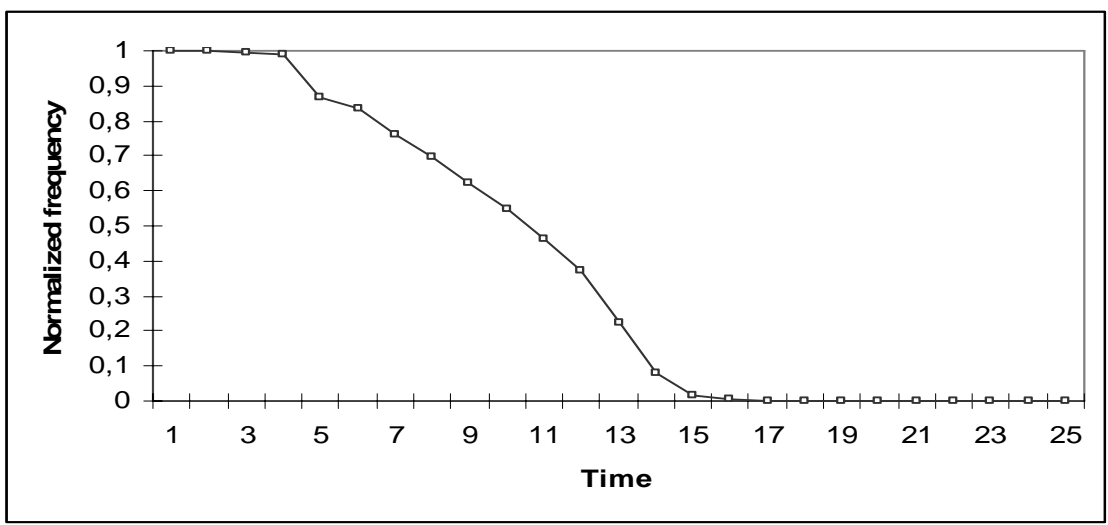

*Fig. 4. Normalized frequency of appearance the singular controls for soft landing example*

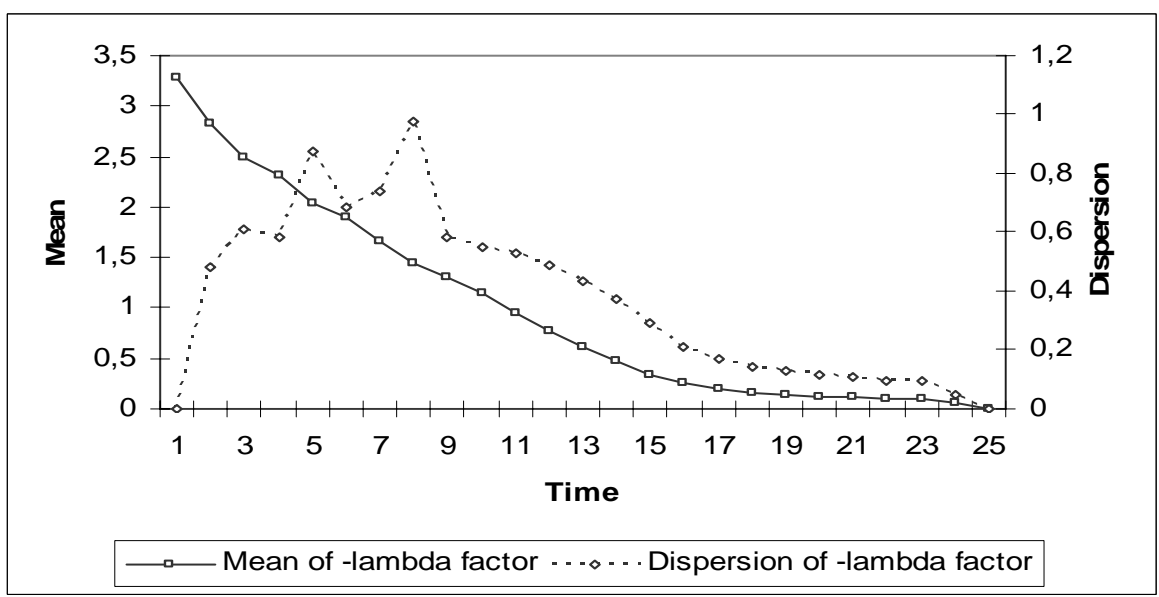

*Fig. 5. Trajectories of mean and dispersion of $-\lambda(k)$ factor*

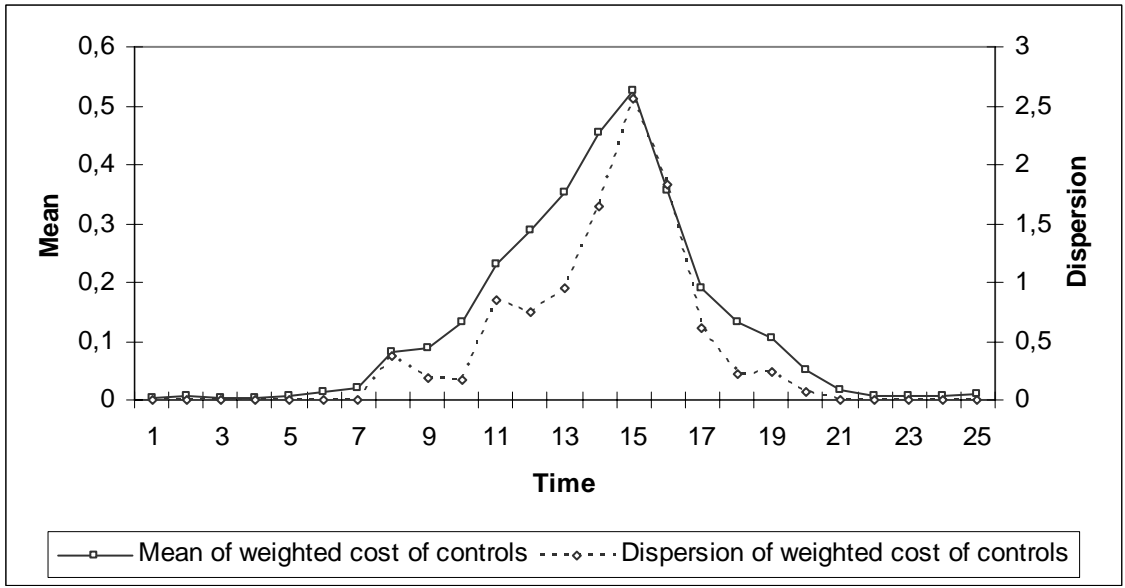

*Fig. 6. Trajectories of mean and dispersion of current weighted control energy*

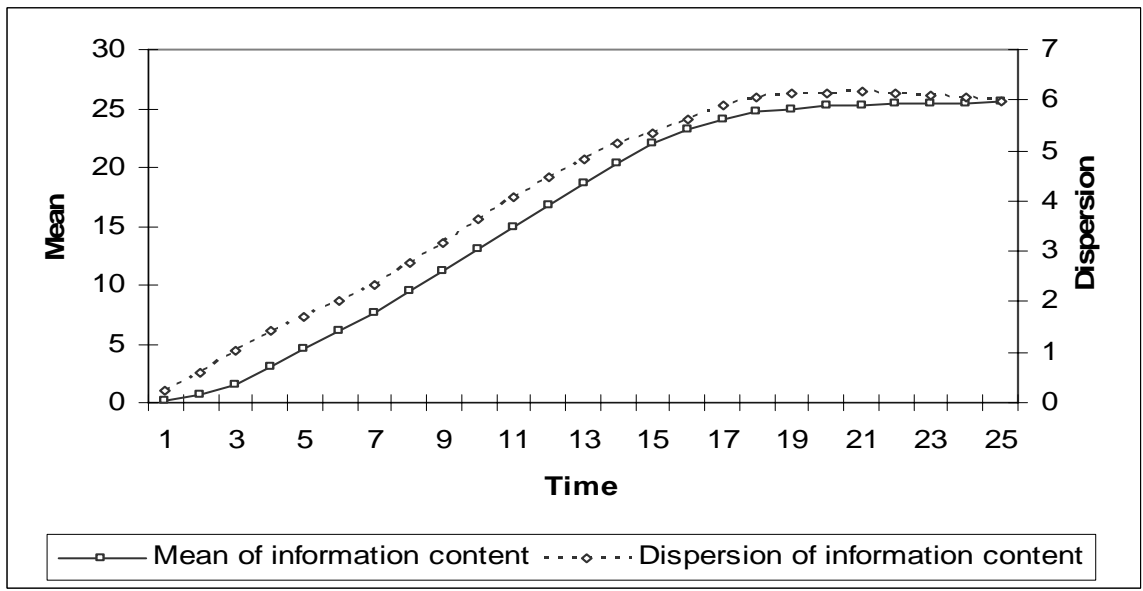

*Fig. 7. Trajectories of mean and dispersion of real cumulative information content about inexact known parameter*

Analysis

Interception problem, regardless that prima facie it may appear that this one rather simpler, requires high level of "precaution", because of the best performances were provided by data processing/control algorithm of type SYS=2/CNT=1. Acceptable performances were showed up by SYS=1/CNT=1 algorithm (better then SYS=2/CNT=0 and SYS=2/CNT=2 variants). It seems that worst results of SYS=0/CNT=0 variant was stipulated by costs surges (and/or miss distance ones), which biased the values of costs mean and dispersion. SYS=2/CNT=2 algorithm provides the best miss distance characteristics, but it seems that in this case occurred some overshoots, which essentially increased the values of cost.

Situation essentially changed in case of solving the soft landing problem, for which significance of "study" factor prevailed over "precaution" one. Best characteristics over all parameters were provided by SYS=2/CNT=2 algorithm, SYS=2/CNT=1 variant provided rather worse performances. Besides the ratings of SYS=1/CNT=1 and SYS=2/CNT=1 algorithms were interchanged, and performances of SYS=0/CNT=0 variant were noticeably improved as well. All above may be considered as additional confirmation of a high importance of "study" factor for effective solving the problems of soft landing-type.

Upgraded algorithm of SYS=2/CNT=2 variant is a most interesting from all points of view, therefore let consider the processes, which occur when this control is applied.

There may be picked out three periods for this algorithm: study (or "turbulent") phase ($0<k\leq5$), transient processes phase ($5<k\leq21$), and proper soft-landing (or "laminar") phase ($21<k\leq25$).

During the first phase an overwhelming majority of applied controls were probe (study) ones of relatively low amplitudes. This phase is characterized by high values of $-\lambda(k)$ factor (or, applying above mentioned analogy, temperatures), and high rate of accumulation of information content. All these point to chaotic type of the processes, which have occurred during this phase, and causes application of "turbulent" term.

At the middle phase behavior of system is determined by transient processes, which occur due to changing of applied controls. Initial interval of this phase is not concerned to overall target of control (solving the soft landing problem), and application of probe controls lead to essential displacements from optimal trajectory. Transition to optimal control stipulated the appearance of intensive control actions, which provides compensation of possible high offsets. Intensiveness of controls supplies saving the high rate of accumulation of information about inexact known parameters. Also there saved the high speed of temperature decreasing (dropping of $-\lambda(k)$ factor).

At the third ("laminar") phase there were provided the solution of proper terminal control problem. This phase is characterized by low control amplitudes and negligible temperatures, as well as stationary state of entropy of inexact known parameters distribution (there is no accumulation of information content).

This model describes the processes, which take place at a micro-level, and characterizes each separate realization of data processing/control algorithm. At a macro-level, if ergodic hypotheses may be applied (averaging over time is equal to one over sets), it is possible to use another model that includes following periods:

- "Childhood" phase (coincide with "turbulent" one) – actions (controls) are chaotic and not goal-seeking (sensible) ones, its intensity are low; there occurs intensive accumulation of information.
- "Youth" phase ($5< k\leq 15$) – actions begin to be goal-seeking, and its intensity and purposefulness is increased; high rate of accumulation of information is saved. Towards the end of this phase intensity of actions is maximal.
- "Ripeness" phase ($15< k\leq 21$) – Majority of actions are goal-seeking ones, and begins decreasing its intensity. At the beginning of the "ripeness" phase rate of accumulation of information begins to drop, and towards the end of this period is sensibly equal to zero.
- "Old age" phase (coincide with "laminar" one) – Actions are fully goal-seeking ones, its intensity is minimal. Rate of accumulation of information is equal to zero.

Of course, this analogy is highly mechanistic, but it seems that usage of such model will provide a rather adequate description of evolution of considered factors for wide class of natural and socioeconomic phenomena/objects.

If summarize all above, the following may be stated:

- Effective solving of optimal stochastic control problems requires subdivision of these ones into two classes: simple and complex. However there not exist evident criteria for selecting these classes now.
- For simple cases (such as interception-type problem) the most effective solutions will be provided by optimal controls for LDS with irreducible PU. If this applicable from effectiveness point of view, algorithm of type SYS=1/CNT=1 is preferred since its realization is most simple. Furthermore, the carried out simulation shows up following property of this option: if value of N was sufficiently high, matrix of feedback gains may have a steady state, i.e. solutions of stabilization-type problems may be realized as some regulator.
- For complex cases (such as soft landing-type problem) additional costs concerning to "study" may be reasonable, and SYS=2/CNT=2 algorithm will be most effective. However this variant may lead to appearance of singular modes, which require application of probe controls for which there are no common-type rules for choosing its parameters. Nevertheless it seems that such mode of optimal control is enough promising and require in-depth study.

## Conclusions

This paper presents common-type solutions for data processing and optimal control problems for LDS with PUs of various types. Proposed solutions possess properties of theoretical clearness and computational effectiveness that is proven by outcomes of carried out numerical simulation. It seems that usage of proposed methods to designing anew and upgrading existing measuring/control systems for various application should essentially improve performances such as reliability, accuracy, stability etc.